\documentclass[sigconf,nonacm]{acmart}

\AtBeginDocument{%
  }

\setcopyright{acmlicensed}
\copyrightyear{2018}
\acmYear{2018}
\acmDOI{XXXXXXX.XXXXXXX}
\acmISBN{978-1-4503-XXXX-X/2018/06}

\usepackage{times}
\usepackage{latexsym}

\usepackage[T1]{fontenc}
\usepackage[T5]{fontenc}

\usepackage[utf8]{inputenc}

\usepackage{multirow}

\usepackage{graphicx}
\usepackage{amsmath}
\usepackage{listings}
\usepackage{tabularx, array}
\usepackage{enumitem}
\usepackage{verbatim}
\usepackage{tcolorbox}
\usepackage{listings}
\usepackage{threeparttable} %
\usepackage{array}        %
\usepackage{url}

\usepackage{booktabs}

\def\eg{{\it e.g.}}

\def\ie{{\it i.e.}}

\begin{document}

\title{MobQA: A Benchmark Dataset for Semantic Understanding of Human Mobility Data through Question Answering}

\author{Hikaru Asano}
\authornote{Work done during an internship at CyberAgent.}
\email{asano-hikaru19@g.ecc.u-tokyo.ac.jp}
\affiliation{%
  \institution{The University of Tokyo \\RIKEN AIP}
  \country{Japan}
  \city{Tokyo}
}

\author{Hiroki Ouchi}
\email{hiroki.ouchi@is.naist.jp}
\affiliation{%
  \institution{Nara Institute of Science and Technology \\RIKEN AIP \\CyberAgent}
  \country{Japan}
  \city{Nara}
}

\author{Akira Kasuga}
\email{kasuga\_akira@cyberagent.co.jp}
\affiliation{%
  \institution{CyberAgent}
  \country{Japan}
  \city{Tokyo}
}

\author{Ryo Yonetani}
\email{yonetani\_ryo@cyberagent.co.jp}
\affiliation{%
  \institution{CyberAgent}
  \country{Japan}
  \city{Tokyo}
}

\begin{abstract}
  This paper presents MobQA, a benchmark dataset designed to evaluate the semantic understanding capabilities of large language models (LLMs) for human mobility data through natural language question answering.
  While existing models excel at predicting human movement patterns, it remains unobvious how much they can interpret the underlying reasons or semantic meaning of those patterns. MobQA provides a comprehensive evaluation framework for LLMs to answer questions about diverse human GPS trajectories spanning daily to weekly granularities. It comprises 5,800 high-quality question-answer pairs across three complementary question types: factual retrieval (precise data extraction), multiple-choice reasoning (semantic inference), and free-form explanation (interpretive description), which all require spatial, temporal, and semantic reasoning. Our evaluation of major LLMs reveals strong performance on factual retrieval but significant limitations in semantic reasoning and explanation question answering, with trajectory length substantially impacting model effectiveness. These findings demonstrate the achievements and limitations of state-of-the-art LLMs for semantic mobility understanding.\footnote{MobQA dataset is available at \url{https://github.com/CyberAgentAILab/mobqa}.}
\end{abstract}




\maketitle

\section{Introduction}

\begin{figure}[h]
    \centering
    \includegraphics[width=\linewidth]{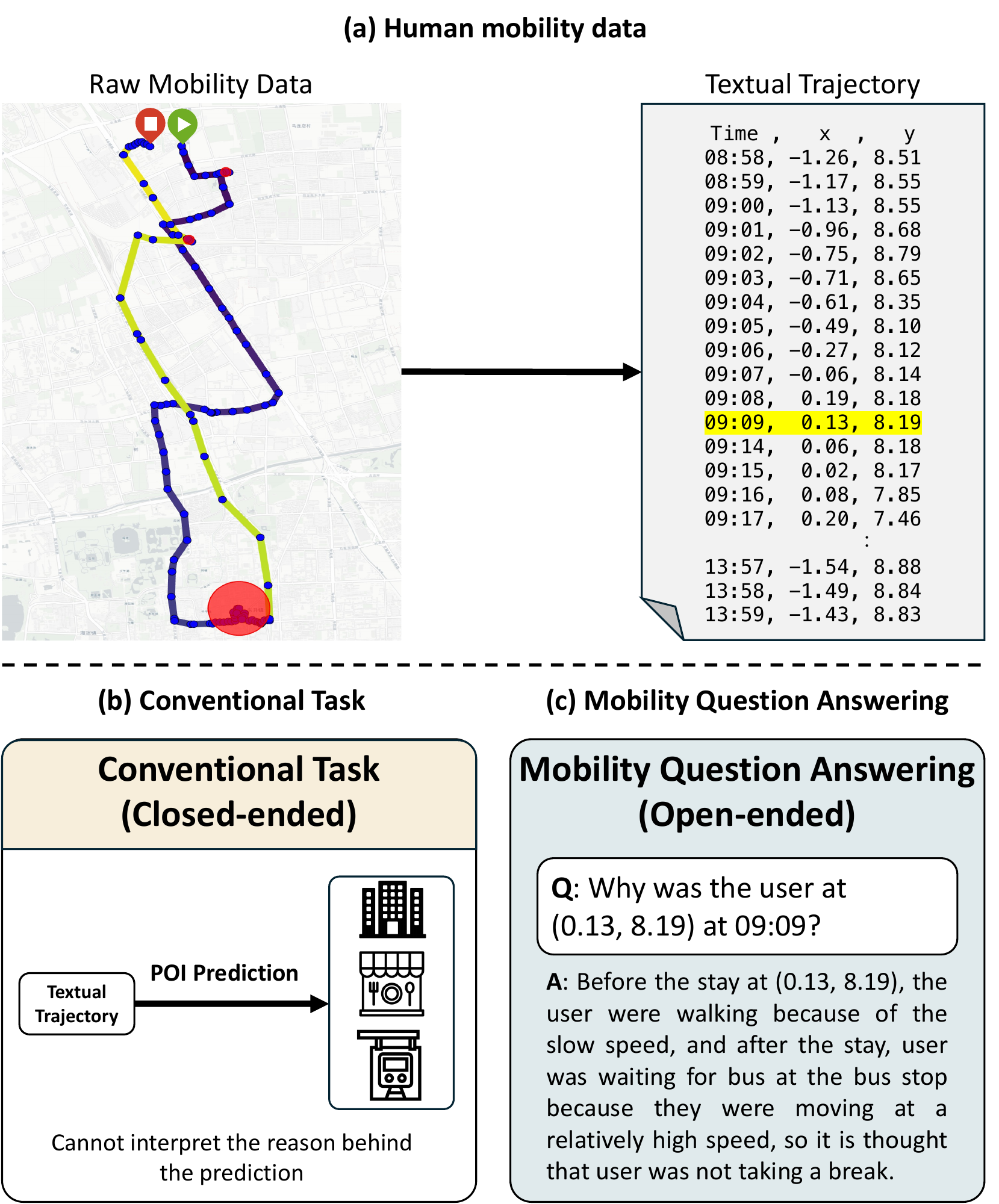}
    \caption{MobQA presents the mobility question answering benchmark. (a) Raw GPS trajectory is anonymized and converted into a textual sequence. (b) Conventional tasks are prediction-oriented and closed-end. (c) Mobility question answering evaluates factual, procedural, and semantic understanding using natural language questions.}
    \label{fig:teaser}
\end{figure}

Modern positioning technologies have enabled large-scale collection of high-resolution human mobility data, but their numerical form makes it difficult for non-experts to interpret or exploit such data~\citep{Parent2013-tx, Albanna2015-qc}. When examining timestamped GPS trajectories like the one shown in Fig.~\ref{fig:teaser}a, it is not straightforward to recognize the mode of transportation, the purpose of the trip, or the reasons behind certain movements. Developing intelligent systems that can understand mobility data from such semantic perspectives would enhance various applications in urban analysis, transportation, and public health~\citep{Shang2011-le, Askarizad2020-ds, Badr2020-oj, Chang2021-pq}, by gaining deeper insights into individual daily activities that go beyond simple statistics on crowd movements.

Prior work has focused primarily on closed-ended, predictive modeling tasks such as next-location prediction and point-of-interest (POI) prediction~\citep{Liu2016-bk, Feng2020-rq, Liang2022-sk, Xu2024-je} (see also Fig.~\ref{fig:teaser}b). These methods are effective at identifying patterns and trends in movement data. However, they typically do not address the underlying reasons behind these patterns, such as user motivations or trip purposes~\citep{Xue2023-dr, Gong2024-tw}.
Indeed, recent literature attributes this limitation to the lack of dedicated datasets and benchmark protocols for addressing the semantic aspects of human mobility data~\citep{Luca2021-ix, Pappalardo2023-wr, Mokbel2024-zr, Zhang2024-oy}.

To address this gap, we present \emph{MobQA}, a new human mobility dataset designed to evaluate intelligent systems through natural language question answering, \ie, \emph{mobility question answering}. Natural language is a flexible and expressive modality for interacting with human mobility data. It allows users to pose open-ended questions about mobility patterns, such as \emph{``Why was the user at this location at this time?''} as illustrated in Fig.~\ref{fig:teaser}c. Furthermore, large language models (LLMs) have recently shown remarkable capabilities in understanding and generating natural language sentences, making them well-suited for interpreting complex mobility data when properly represented as text.

The MobQA Dataset is built upon the Geolife GPS dataset~\citep{Zheng2008-gf,Zheng2010-lk,Zheng2011-mg} and comprises 5,800 high-quality question-answer pairs for human GPS trajectories spanning daily and weekly granularities. The questions fall into three complementary types: 2,000 factual retrieval questions for precise data extraction, 2,000 multiple-choice questions for semantic inference, and 1,800 free-form questions for interpretive explanation. We also establish comprehensive evaluation protocols that use accuracy metrics for factual and multiple-choice tasks and adopt an LLM-as-a-judge framework~\citep{Zheng2023-au, Gu2024-kt} to assess the faithfulness and informativeness of free-form answers, following established evaluation practices~\citep{Sai2022-kr, Adlakha2024-tl}.

We conducted extensive experiments with various LLMs, including GPT-4o series, Gemini 1.5/2.0, o3-mini, and ten open-source models from the Llama 3 series~\citep{Grattafiori2024-ga}, Qwen 3~\citep{Yang2025-yi}, and DeepSeek-R1 series~\citep{DeepSeek-AI2025-vk}, to evaluate their performance on mobility question answering tasks. Our results reveal that while these models achieve strong performance on factual retrieval question answering, they demonstrate significant limitations in semantic reasoning and explanation tasks, with performance substantially degrading for longer trajectory sequences. We believe that the MobQA dataset will encourage the research community to develop more explainable and accessible mobility understanding technologies and applications.

\section{Mobility Question Answering}
\label{sec:preliminaries}

\subsection{Task Description}
\label{subsec:mobility_qa}
We address the task of mobility question answering. This task challenges LLMs to reason over human mobility data and answer natural language questions that require spatial, temporal, and semantic understanding of movement patterns. Formally, given a question $Q$ about a user's movement trajectory $T$, the goal is to generate a natural language answer $A$:
\begin{equation}\label{eq:qa}
(Q, T) \rightarrow A.
\end{equation}
The trajectory $T$ serves as the primary data source, represented as a textual sequence of time-stamped coordinates as illustrated in Fig.~\ref{fig:teaser}a. LLMs need to infer spatiotemporal patterns and semantic cues \emph{solely from this numerical data sequence} $T$ to produce accurate, context-rich answers $A$ to the question $Q$.

\begin{figure*}[t]
    \centering
    \includegraphics[width=\linewidth]{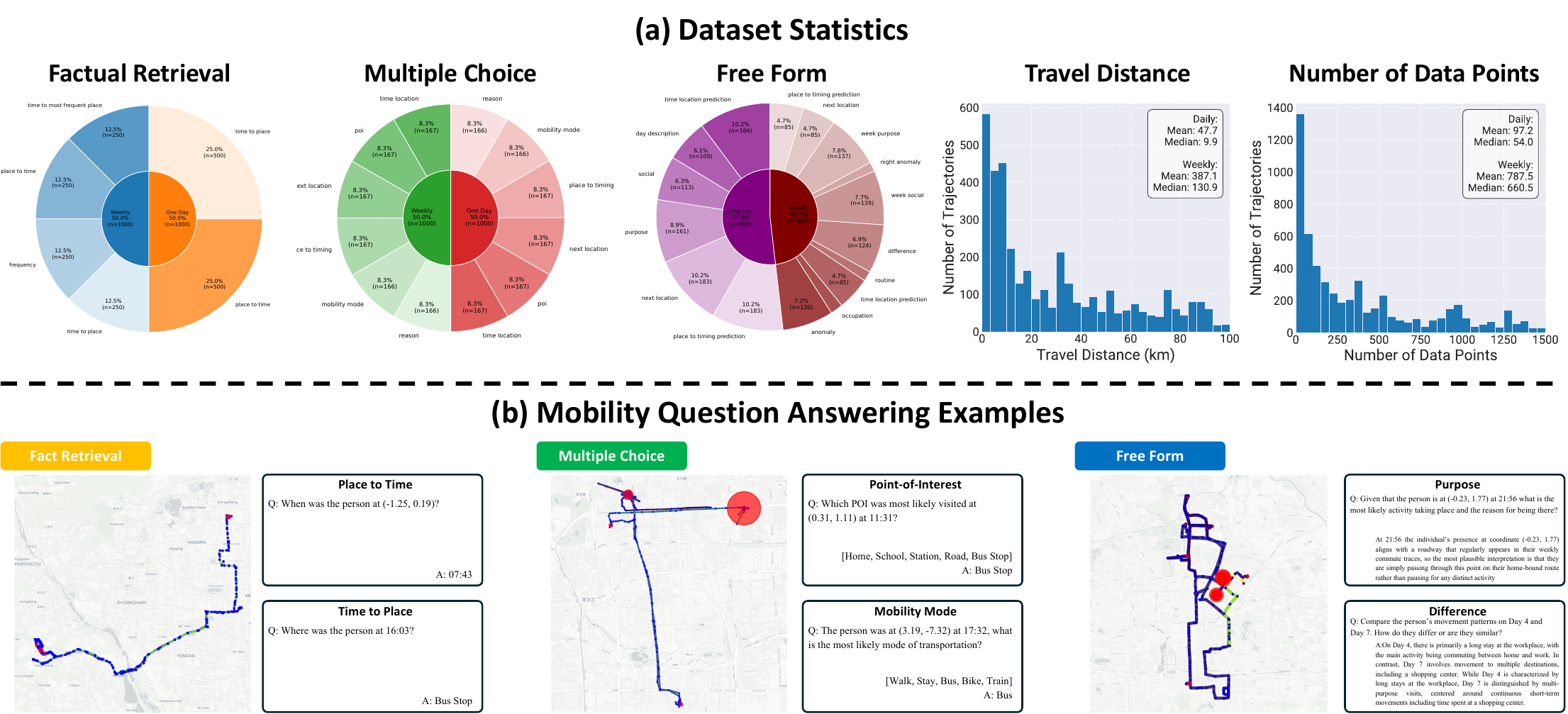}
    \caption{(a) Distribution of questions across the three categories: factual retrieval, multiple choice, and free form, as well as the distributions of trajectory length (number of data points) and travel distance in the MobQA dataset. (b) Representative examples of questions and answers for each category, along with the corresponding trajectories.}
    \label{fig:mobility_qa_overview} %
\end{figure*}

\subsection{Trajectories}
Human trajectories are typically long sequences; collecting GPS data every minute yields 1,440 points for a single day. Even after removing stationary moments to reduce redundancy, daily trajectories can still comprise several hundred points. This poses challenges for current LLMs that can struggle with such long sequences of numerical data~\citep{Kim2024-us, Liu2024-vc, Yoon2024-cy}. To systematically examine the impact of input sequence lengths on LLM question answering performance, we consider two trajectory granularities:
\begin{itemize}[itemsep=1.5pt, topsep=5pt, leftmargin=1.2em, labelsep=0.6em, after=\vspace{0.35\baselineskip}]
\item \textbf{Daily Trajectory}: A single 24-hour period, highlighting intraday behaviors such as commuting, evening outings, and various other short-term activities.
\item \textbf{Weekly Trajectory}: A seven-day sequence providing richer behavioral context, revealing weekly habits, recurring patterns, and anomalies.
\end{itemize}
While weekly trajectories provide more behavioral evidence for richer inferences, they include more data points. Evaluating models on questions with those two granularities allows us to quantify the trade-off between contextual richness and sequence length.

\subsection{Questions}

To evaluate LLMs across different reasoning demands, we design three complementary question types inspired by existing benchmarks~\citep{Baradaran2022-ir, Zhong2022-st, Fang2024-hj}: \emph{factual retrieval}, \emph{multiple choice}, and \emph{free form}. Fig.~\ref{fig:mobility_qa_overview} shows statistics and examples for each question.
\begin{itemize}[itemsep=1.5pt, topsep=5pt, leftmargin=1.2em, labelsep=0.6em, after=\vspace{0.35\baselineskip}]
    \item \textbf{Factual retrieval questions} evaluate an LLM's ability to \emph{locate and extract verifiable facts} from a trajectory by focusing on specific data points. A key challenge is pinpointing the relevant data point in a long sequence, as LLMs must identify specific information among many similar entries. The input-output interface follows Eq.~\eqref{eq:qa}; the answer should be a precise value (\eg, coordinates, times, counts) without complex reasoning. Examples include: \emph{``Where is the user at 10:15?''}
    \item \textbf{Multiple-choice questions} ask LLMs to \emph{infer hidden patterns or semantic attributes}, such as mobility mode or the next POI, from numerical trajectories. The main challenge is extracting key information from the numerical data and identifying important features. Each instance extends the standard $(Q,T)$ pair with five candidate answers (one correct and four distractors), requiring the LLMs to choose the correct one. For example, question is \emph{``What is the mobility mode at 10:15 at (-0.15, 4.91)?''}, and the candidate is \texttt{[Walk, Stay, Bus, Bike, Train]}
    \item \textbf{Free-form questions} require LLMs to generate \emph{descriptive or explanatory interpretations} from the trajectory. Models must abstract complex patterns, detect anomalies, and infer underlying intentions. The output is unconstrained text following the interface in Eq.~\eqref{eq:qa}. Example: \emph{``Summarize Day 7's mobility and explain any deviations from the weekly routine.''}
\end{itemize}
Consequently, our benchmark offers a comprehensive assessment with tasks ranging from basic data retrieval to complex spatiotemporal reasoning and description generation.

\subsection{Evaluation Methodology}
\label{subsec:evaluation_methodology}
We evaluate the performance of LLMs on mobility question answering using metrics tailored to each question type.
For factual retrieval and multiple-choice questions with precise answers, we use the \emph{accuracy} metric, which calculates if the answer exactly matches the ground truth values or choices. On the other hand, we adopt \emph{faithfulness} and \emph{informativeness} metrics~\citep{Sai2022-kr, Adlakha2024-tl} for the free-form questions because they are subjective and require evaluating both factual accuracy and semantic richness:
\begin{itemize}[itemsep=1.5pt, topsep=5pt, leftmargin=1.2em, labelsep=0.6em, after=\vspace{0.35\baselineskip}]
    \item \textbf{Faithfulness} assesses how accurately the generated answers reflect factual details (\eg, coordinates, timestamps) from the trajectory, ensuring alignment and no contradictions.
    \item \textbf{Informativeness} evaluates the completeness and richness of the response, verifying that it comprehensively addresses the question by incorporating relevant movement patterns and inferred semantic insights.
\end{itemize}
To evaluate free-form responses, we employ an \emph{LLM-as-a-judge framework}~\citep{Zheng2023-au, Gu2024-kt} that uses powerful LLMs to score generated responses (see Sec.~\ref{sec:experimental_setup} for more details.)

\begin{figure*}[t]
    \centering
    \includegraphics[width=\linewidth]{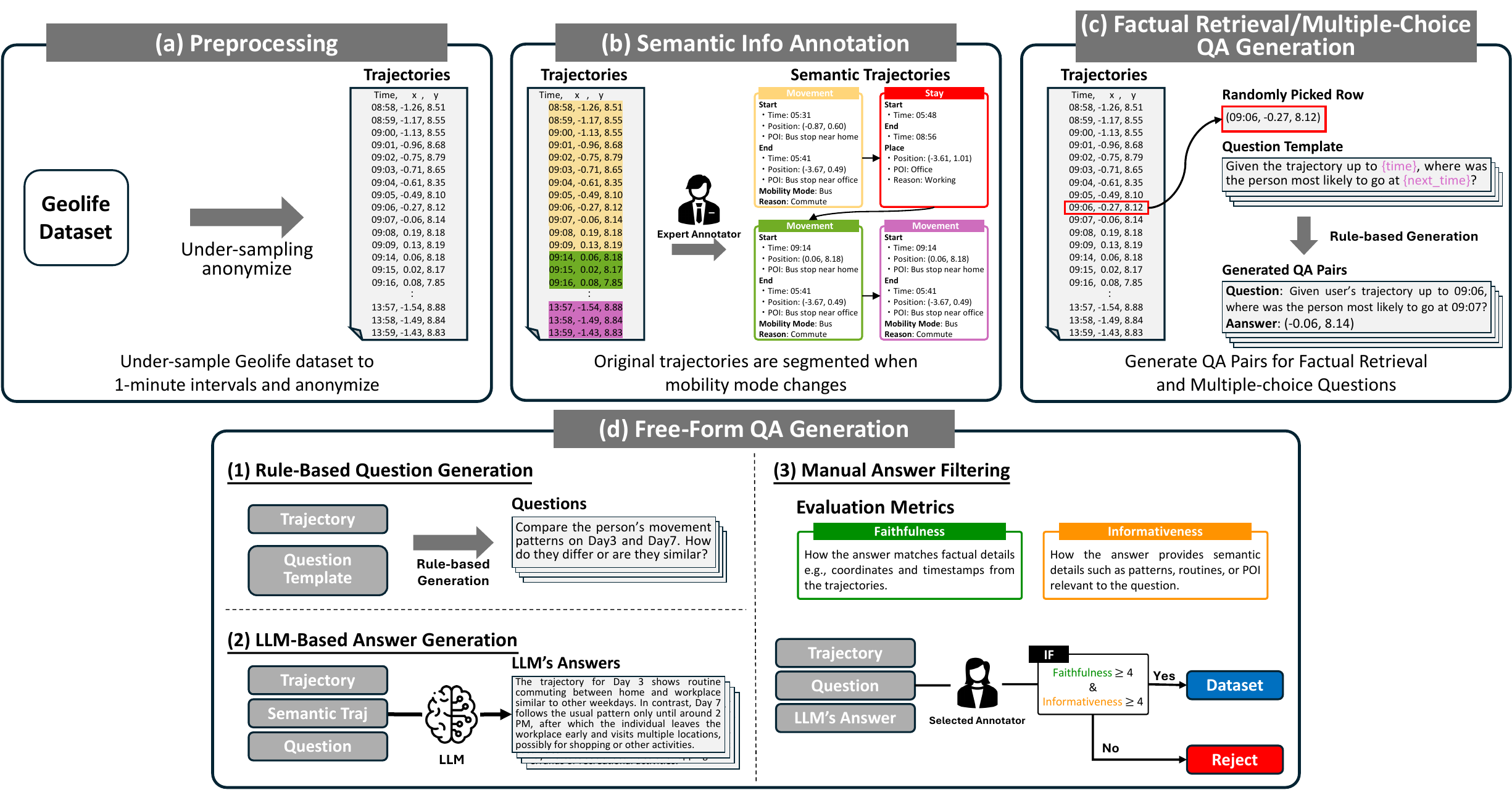}
    \caption{Overview of the MobQA dataset construction process, illustrating the stages of data preprocessing, semantic trajectory construction, question generation, answer generation, and answer filtering.}
    \label{fig:construction_process}
\end{figure*}

\section{MobQA Dataset}
\label{sec:dataset_construction}
This section details our data construction pipeline (see also Fig.~\ref{fig:construction_process}). The proposed MobQA dataset comprises 5,800 high-quality question-answer pairs across three complementary question types introduced in the previous section: 2,000 factual retrieval questions for precise data extraction, 2,000 multiple-choice questions for semantic inference, and 1,800 free-form questions for interpretive explanation.

\subsection{Constructing Trajectory Data}
\label{subsec:trajectory_data}

\paragraph{Mobility Dataset.}
We build the MobQA dataset using the Geolife GPS dataset~\citep{Zheng2008-gf,Zheng2010-lk,Zheng2011-mg}, selected after systematically reviewing publicly available human mobility datasets~\citep{Calabrese2015-io, Zhao2016-ci, Luca2021-ix}. Geolife offers a unique combination of high spatiotemporal resolution GPS trajectories recorded at frequent intervals, verified multi-modal transportation labels across 11 transport modes (\eg, walking, biking, subway, car), and diverse mobility behaviors spanning commutes, leisure, and long-distance trips. With longitudinal coverage exceeding 5 years for 182 users, the dataset enables both fine-grained movement analysis and temporal pattern identification. These characteristics make Geolife the only publicly available dataset suitable for generating semantically rich question-answer pairs that demand understanding of complex spatiotemporal movement patterns, in contrast to other alternative datasets. For example, check-in logs capture only discrete location events~\citep{Yang2015-wj, Cho2011-dz}, vehicle traces cover single transport modes~\citep{Yuan2010-lb}, and cellular data lacks spatial precision~\citep{Blondel2012-ra} (See Appendix~\ref{sec:base_dataset_justification} for further details regarding our selection process.)

\paragraph{Preprocessing.}
We create the input textual trajectory, $T$ in Eq.~\ref{eq:qa}, by converting raw GPS data into a structured format suitable for LLMs. From the Geolife dataset, we retain 69 participants with user-reported mobility-mode labels from the original 182 users. Raw data consists of spatiotemporal movement records represented as ordered triples of geographic coordinates and timestamps:
\begin{equation}
\bigl\{ (x_i, y_i, t_i) \mid i = 1, \ldots, n \bigr\},
\label{eq:raw_gps}
\end{equation}
where $(x_i, y_i)$ denote the latitude and longitude coordinates of the $i$-th GPS observation and $t_i$ represents the corresponding timestamp. We downsample these logs to 1-minute intervals and merge stationary periods into single events, reducing data volume while preserving sufficient temporal resolution.
Despite these preprocessing steps, our dataset features long sequences; daily trajectories average 97.2 data points (median: 54.0), while weekly trajectories average 787.5 data points (median: 660.5), as shown in Fig.~\ref{fig:mobility_qa_overview}, reflecting a heavy-tailed distribution in movement intensity across users.

\paragraph{Privacy-Preserving Processing.}
To enhance privacy protection and standardize trajectory coordinate, all geographic coordinates are transformed into anonymized offsets $(x_i', y_i')$ relative to a randomly selected user-specific reference point $(x_{\text{ref}}, y_{\text{ref}})$, computed as $(x_i', y_i') = (x_i - x_{\text{ref}}, y_i - y_{\text{ref}})$. This coordinate anonymization removes personally identifiable location information while preserving the essential geometric structure and relative spatial relationships necessary for trajectory analysis.

\paragraph{Textual Trajectory Representation.}
Each trajectory is converted into plain text sequences to be compatible with LLM processing. Specifically, we represent each data point as a tuple of the form $(\text{hh:mm}, x_i', y_i')$, where $\text{hh:mm}$ denotes the time in hours and minutes, and $x_i', y_i'$ are the anonymized coordinates rounded to two decimal places (approximately 10-meter resolution). The resulting representation is a sequence of such tuples, separated by newline characters (\texttt{\textbackslash n}), as shown in the following example:
\begin{center}
\texttt{(09:02, -0.75, 8.79)\textbackslash n(09:03, -0.71, 8.65)\textbackslash n\ldots(14:00, -1.38, 8.78)}.
\end{center}
For a complete illustration of a daily trajectory represented in this textual format, refer to Fig.~\ref{fig:trajectory_example} in Appendix~\ref{sec:full_dataset_result}

\subsection{Annotating Semantic Information}
\label{subsec:semantic_annotation}
We further enrich a subset of the trajectory data (eight consecutive days from 19 of the 69 retained users, selected to maximize geographic coverage, routine diversity, and transport mode variety) with manual annotations to generate question-answer pairs that require detailed semantic understanding of mobility data.

Three professional annotators specialized in mobility-data annotation performed the labeling using a dedicated interactive annotation tool we developed to facilitate minute-level precision (see Appendix~\ref{subsec:annotation_tool} for more details).
As shown in Fig~\ref{fig:construction_process}b, each daily trajectory is segmented whenever (1) the \emph{mobility mode} changes (\eg, from walking to bus) or (2) a \emph{stay phase} (\eg, at home or work) begins or ends. The annotators then label every resulting segment with:
\begin{itemize}[itemsep=1.5pt, topsep=5pt, leftmargin=1.2em, labelsep=0.6em, after=\vspace{0.35\baselineskip}]
\item \textbf{Point of Interest (POI):} semantic location types (\eg, home, workplace, shopping center),
\item \textbf{Mobility Mode:} transportation methods (\eg, walking, bus, subway, car),\footnote{We re-annotated mobility modes here because, although the Geolife dataset originally includes self-reported mobility modes, they are often temporally inaccurate (off by up to one minute) or sometimes missing.} and
\item \textbf{Reason:} underlying purposes for the movement or stay (\eg, commuting, leisure, shopping).
\end{itemize}
All the annotators were thoroughly briefed on the study purpose and provided informed consent for their annotations to be included in the publicly available dataset. The annotation protocol involved two structured practice phases using pre-annotated exemplar trajectories.

The lead author performed final validation for quality assurance of all annotations to ensure consistency in mobility modes, precise segment boundary alignment, and completeness. This process required approximately 26 minutes per daily trajectory and 60 hours total, yielding high-fidelity semantic annotations covering 11 verified transport modes including walking, biking, subway, and car, spanning diverse mobility behaviors from commutes and leisure activities to long-distance trips. See Appendix~\ref{sec:semantic_traj} for details.

\subsection{Creating Question-Answer Pairs}
\label{sed:qa_pair_generation}
\paragraph{Factual Retrieval Questions}

Factual retrieval questions assess the ability of LLMs to extract precise information directly from trajectory data. These questions require exact data retrieval without interpretation, such as determining a person's location at a specific time or counting visit frequencies.
Specifically, we employ a template-driven pipeline using predefined question templates:
\begin{itemize}[itemsep=1.5pt, topsep=5pt, leftmargin=1.2em, labelsep=0.6em, after=\vspace{0.35\baselineskip}]
\item \texttt{time\_to\_place}: ``Where was the person at \{time\}?''
\item \texttt{place\_to\_time}: ``When was the person at \{place\}?''
\item \texttt{frequency}: ``How many times did the person visit the location exactly at coordinates \{place\}?''
\end{itemize}
Our pipeline involves: (1) uniformly sampling a trajectory suitable for the target question type, (2) selecting a single data point from the sampled trajectory, and (3) populating template placeholders with extracted attributes to formulate the final question. Answers are produced deterministically by directly copying requisite fields or applying simple rule-based operations.

\paragraph{Multiple-Choice Questions}

Multiple-choice questions evaluate semantic inference capabilities, requiring LLMs to understand context and make reasoned selections from provided options. These questions assess understanding of mobility modes, points of interest, purposes, and predictive reasoning. Similar to factual retrieval, we use templates for question generation. 
\begin{itemize}[itemsep=1.5pt, topsep=5pt, leftmargin=1.2em, labelsep=0.6em, after=\vspace{0.35\baselineskip}]
\item \texttt{poi}: ``Given that the person was at \{place\} at \{time\}, which point of interest did they most likely visit?''
\item \texttt{mobility\_mode}: ``The person was at \{place\} at \{time\}, what is the most likely mode of transportation?''
\item \texttt{next\_location}: ``Given the trajectory up to \{time\}, where was the person most likely to go at \{next\_time\}?''
\end{itemize}
We sample questions from the full corpus of trajectories if they can be answered from raw GPS traces. If the questions otherwise require the inference of semantics (\eg, mobility mode, activity purpose), we use the semantically enriched data subset created in Sec.~\ref{subsec:semantic_annotation} to ensure that the necessary semantic information is available for verifying the answers. In addition to correct answers, we crafted four plausible yet unambiguously incorrect distractors for each question, Numeric attributes for the distractors are drawn from the same trajectory, and semantic attributes for them are selected from the established vocabulary during semantic trajectory annotation.

\paragraph{Free-Form Questions}

Free-form questions require comprehensive analysis and interpretive explanation, assessing the LLMs' ability to provide rich contextual descriptions of mobility patterns, compare different time periods, and infer underlying reasons for movements. We again use template generation and sample questions from semantically enriched trajectories:
\begin{itemize}[itemsep=1.5pt, topsep=5pt, leftmargin=1.2em, labelsep=0.6em, after=\vspace{0.35\baselineskip}]
\item \texttt{routine}: ``Analyze the weekly trajectory to identify and summarize the individual's regular movement patterns and routine activities.''
\item \texttt{difference}: ``Compare the person's movement patterns on \{day1\} and \{day2\}. How do they differ or are they similar?''
\item \texttt{anomaly}: ``Evaluate whether the mobility pattern on \{day\} deviates from the individual's established routine.''
\end{itemize}
Unlike factual or multiple-choice questions, creating answers for free-form questions cannot be done by predefined rules alone. As shown in the Fig~\ref{fig:construction_process}d, we first leverage GPT-4o to generate fluent, evidentially grounded responses. Specifically, we feed the question tuple $(Q, T)$ as well as semantic information annotated in Sec.~\ref{subsec:semantic_annotation} to GPT-4o to obtain candidate answers.

For manual filtering, we first recruited nine annotators from a crowd-sourcing platform for a pilot task. To ensure the quality of evaluation, we retained the three best‑performing annotators for the main filtering phase. They evaluated the candidate answers by scoring their faithfulness and informativeness (Sec.~\ref{subsec:evaluation_methodology}), and only answers with scores exceeding 4 for both criteria were accepted into the dataset.
See Appendix~\ref{sec:manual_answer_filtering} for manual filtering process details.

\section{Experiments}
\label{sec:experiment}
With the MobQA dataset, we conducted a series of experiments to evaluate the performance of various state-of-the-art LLMs on mobility question answering tasks. 
In what follows, we first evaluate closed-source LLMs in Sec.~\ref{sec:closed_models} and open-sourced models in Sec.~\ref{sec:open_models}. We then investigate how various factors, including trajectory lengths and semantic information, can impact the mobility question performances in Sec.~\ref{sec:traj_length_effect} and Sec.~\ref{sec:semantic_info_effect}.

\subsection{Experimental Setup}
\label{sec:experimental_setup}
\paragraph{Data Partitioning} To ensure fair comparison between open-source and closed-source models, we partitioned the MobQA dataset (5,800 samples) into training (80\%) and test (20\%) sets. The training data were used exclusively for fine-tuning open-source models, while all evaluations were conducted on the held-out test set.

\paragraph{Memory Constraints} Due to GPU memory limitations, our evaluation was restricted to trajectories containing fewer than 1,000 data points.\footnote{Although the precise token count depends on numerical tokenization, a trajectory with 1,000 points typically corresponds to approximately 20,000 tokens including prompt.} All experiments were performed using an NVIDIA A100 GPU. Complete results for closed-source models on the full dataset are provided in the Tab.~\ref{tab:full_dataset_result} in the appendix.

\paragraph{Prompting Strategy} To elicit responses from each model, we designed a structured prompt shown in Fig.~\ref{fig:factual_retrieval_prompt}, containing five components: (1) mobility data description, (2) task-specific instructions, (3) textual trajectory representation, (4) target question, and (5) trajectory data reference.
\paragraph{LLM-as-a-Judge}
To select an appropriate evaluation model for LLM-as-a-judge framework as described in Sec.~\ref{subsec:evaluation_methodology}, we compared several candidate judge models and chose Gemini-2.0-Flash that demonstrated a strong correlation with human ratings across different settings as shown in Tab.~\ref{tab:correlation}.
To examine potential self-enhancement bias~\citep{Zheng2023-au, Dai2024-cr, Ye2025-yh}, where LLMs favor responses from their own model family, we re-scored the answers with GPT-4o. As shown in Tab.~\ref{tab:eval_comparison}, the ratings from both models were highly consistent, suggesting self-enhancement bias was limited in our task.
Additionally, our pilot experiment confirmed that conventional automatic metrics, such as BLEU, ROUGE, and BERTScore, showed poor alignment with human evaluations (See Appendix~\ref{sec:evaluation_methodology_details} for details).

\begin{table}[t]
  \centering
  \caption{Correlation coefficients (Pearson and Spearman) between LLM and human evaluations for faithfulness and informativeness metrics}
  \label{tab:correlation}
  \footnotesize
  {\setlength{\tabcolsep}{4pt}%
  \begin{tabular}{lcccc}
    \toprule
    \multirow{2}{*}{Model} & \multicolumn{2}{c}{Faithfulness} & \multicolumn{2}{c}{Informativeness} \\
    \cmidrule(lr){2-3} \cmidrule(lr){4-5}
                           & Daily & Weekly & Daily & Weekly \\
    \midrule
    GPT-4o-mini           & 0.502 / \textbf{0.562} & 0.278 / 0.349 & 0.554 / 0.605 & 0.402 / 0.425 \\
    GPT-4o                & \textbf{0.507} / 0.524 & 0.401 / 0.492 & 0.521 / 0.511 & 0.513 / 0.537 \\
    Gemini 1.5            & 0.370 / 0.378 & \textbf{0.513} / \textbf{0.548} & 0.544 / 0.593 & 0.502 / 0.521 \\
    Gemini 2.0            & 0.368 / 0.401 & 0.494 / 0.535 & \textbf{0.584} / \textbf{0.608} & \textbf{0.577} / \textbf{0.596} \\
    \bottomrule
  \end{tabular}
  }%
\end{table}

\subsection{Evaluation of Closed-source LLMs}
\label{sec:closed_models}
We benchmark five recent LLMs: GPT-4o-mini-2024-07-18, GPT-4o-2024-11-20, o3-mini-2025-01-31, Gemini-1.5-Flash-002, and Gemini-2.0-Flash-001. All models were accessed through their public API. The details of models and prompts used for generation are detailed in Appendix~\ref{sec:experiment_detail}.

Tab.~\ref{tab:combined_results} reveals distinct patterns of performances across different tasks and data granularities. For factual retrieval, the Gemini models demonstrated superior accuracy, with Gemini-2.0-Flash achieving a near-perfect score of 0.991 on the daily trajectories. However, we observed consistent performance degradation across all models when processing the longer weekly trajectories. This finding highlights a key challenge for current models in maintaining factual recall as sequence length increases, which we will analyze further in Sec.~\ref{sec:traj_length_effect}. In contrast, results on multiple-choice questions exhibited a different trend. Model performances were largely comparable between the daily and weekly trajectory inputs, with only minor differences observed across most models, suggesting the task dependency for the impact of input lengths.

\begin{table}[t]
  \centering
  \caption{Performance comparisons of closed-source LLMs}
  \label{tab:combined_results}
  \scriptsize
  {\setlength{\tabcolsep}{3.1pt}%
  \begin{tabular}{lcccccccc}
    \toprule
    \multirow{2}{*}{Model} & \multicolumn{2}{c}{Factual Retrieval} & \multicolumn{2}{c}{Multiple-Choice} & \multicolumn{2}{c}{Faithfulness} & \multicolumn{2}{c}{Informativeness} \\
    \cmidrule(lr){2-3} \cmidrule(lr){4-5} \cmidrule(lr){6-7} \cmidrule(lr){8-9}
           & Daily & Weekly & Daily & Weekly & Daily & Weekly & Daily & Weekly \\
    \midrule
    GPT-4o-mini & 0.690 & 0.462 & 0.294 & 0.310 & 2.00 & 2.12 & 1.96 & 2.13 \\
    GPT-4o      & 0.812 & 0.550 & 0.413 & 0.457 & 2.09 & 2.35 & 2.10 & 2.43 \\
    o3-mini     & 0.824 & 0.673 & 0.472 & \textbf{0.505} & 2.45 & 2.78 & 2.42 & 2.86 \\
    Gemini 1.5  & 0.963 & \textbf{0.880} & \textbf{0.490} & 0.449 & \textbf{3.15} & 2.86 & \textbf{3.13} & 2.97 \\
    Gemini 2.0  & \textbf{0.991} & 0.858 & 0.403 & 0.405 & 2.96 & \textbf{3.10} & 2.90 & \textbf{3.23} \\
    \bottomrule
  \end{tabular}
  }%
\end{table}

The free-form questions proved to be the most challenging for all models, with faithfulness and informativeness scores typically ranging between 2.0 and 3.0 out of 5. Notably, while o3-mini is recognized for its state-of-the-art reasoning capabilities, its performance was limited across all tasks. This outcome aligns with recent observations that general-purpose reasoning mechanisms, such as chain-of-thought, may not readily translate to domains requiring nuanced interpretation of dense, sequential data like time series~\citep{Zhou2025-zo}. These results show that generating high-quality, faithful, and informative free-form responses from mobility data remains a significant open challenge for current LLMs. See Appendix~\ref{sec:free_form_examples} for qualitative examples of model outputs.

\subsection{Evaluation of Open-source Models}
\label{sec:open_models}

To ensure the reproducibility of our experiments and analyze how performance varies with model size, we benchmarked ten open-source models from the Llama 3.x~\citep{Grattafiori2024-ga}, Qwen 3~\citep{Yang2025-yi}, and DeepSeek-R1 distilled model series~\citep{DeepSeek-AI2025-vk}. Furthermore, to evaluate the performance of LLMs specialized for mobility data
we finetuned on four of these models via Supervised Fine-Tuning (SFT)~\citep{Ouyang2022-gr, Wang2023-st}. See Appendix~\ref{sec:experiment_detail} for the details of the base models and the SFT process.

\begin{table}[t]
  \centering
  \caption{Performance comparisons of open-source LLMs and their fine-tuned versions}
  \label{tab:open_model_results}
  \scriptsize
  {\setlength{\tabcolsep}{2.8pt}%
  \begin{tabular}{lcccccccc}
    \toprule
    \multirow{2}{*}{Model} & \multicolumn{2}{c}{Factual Retrieval} & \multicolumn{2}{c}{Multiple-Choice} & \multicolumn{2}{c}{Faithfulness} & \multicolumn{2}{c}{Informativeness} \\
    \cmidrule(lr){2-3} \cmidrule(lr){4-5} \cmidrule(lr){6-7} \cmidrule(lr){8-9}
           & Daily & Weekly & Daily & Weekly & Daily & Weekly & Daily & Weekly \\
    \midrule
    \multicolumn{9}{l}{\textit{Open Source Models}} \\
    Llama-3.2-1B & 0.000 & 0.000 & 0.045 & 0.019 & 1.00 & 1.15 & 1.00 & 1.15 \\
    Llama-3.2-3B & 0.615 & 0.168 & 0.210 & 0.087 & 1.01 & 1.48 & 1.01 & 1.53 \\
    Llama-3.1-8B & 0.840 & \textbf{0.679} & 0.295 & 0.237 & 1.01 & 1.25 & 1.02 & 1.40 \\
    Qwen3-1.7B & 0.110 & 0.007 & 0.045 & 0.048 & 1.07 & 1.09 & 1.07 & 1.09 \\
    Qwen3-4B & 0.815 & 0.350 & \textbf{0.560} & 0.375 & 1.47 & \textbf{1.94} & 1.48 & 1.96 \\
    Qwen3-8B & \textbf{0.920} & 0.482 & 0.495 & 0.319 & 1.22 & 1.35 & 1.26 & 1.36 \\
    Qwen3-14B & 0.835 & 0.460 & 0.265 & 0.237 & 1.90 & 1.31 & 1.90 & 1.31 \\
    R1-1.5B & 0.090 & 0.000 & 0.015 & 0.013 & 1.28 & 1.49 & 1.28 & 1.50 \\
    R1-7B & 0.570 & 0.015 & 0.030 & 0.000 & 1.79 & 1.37 & 1.78 & 1.35 \\
    R1-14B & 0.875 & 0.387 & 0.050 & 0.150 & \textbf{2.36} & 1.68 & \textbf{2.33} & 1.67 \\
    \midrule
    \multicolumn{9}{l}{\textit{Fine-tuned Models}} \\
    Llama-3.2-1B (FT) & 0.485 & 0.241 & 0.435 & 0.444 & 2.11 & 1.69 & 2.16 & 1.81 \\
    Llama-3.2-3B (FT) & 0.730 & 0.328 & 0.505 & 0.562 & 2.10 & 1.61 & 2.18 & 1.67 \\
    Qwen3-1.7B (FT) & 0.650 & 0.307 & 0.510 & 0.494 & 2.18 & 1.50 & \textbf{2.33} & 1.64 \\
    Qwen3-4B (FT) & 0.830 & 0.460 & \textbf{0.560} & \textbf{0.606} & 2.14 & 1.62 & 2.23 & \textbf{1.84} \\
    \bottomrule
  \end{tabular}
  }%
\end{table}

The evaluation results in Tab.~\ref{tab:open_model_results} indicate a general positive correlation between model size and performance. Smaller models, particularly in the 1B class, struggled with benchmark tasks. While larger models demonstrated improved performance, their capabilities remained limited, especially when handling weekly trajectories that require long-term contextual understanding. This difficulty is particularly evident in multiple-choice questions, where most open models performed worse on weekly trajectories compared to more capable closed models, suggesting the less effectiveness of those models for extracting meaningful information from longer trajectories. The model series showed varying strengths. For instance, the Qwen series excelled at the multiple-choice tasks, while Qwen3-8B and Llama-3.1-8B are good at daily and weekly factual retrieval tasks, respectively. The R1 series demonstrated best for the daily free-form tasks. This performance diversity highlights our benchmark's effectiveness in assessing multifaceted model capabilities.

\begin{figure*}[t]
  \centering
  \includegraphics[width=\linewidth]{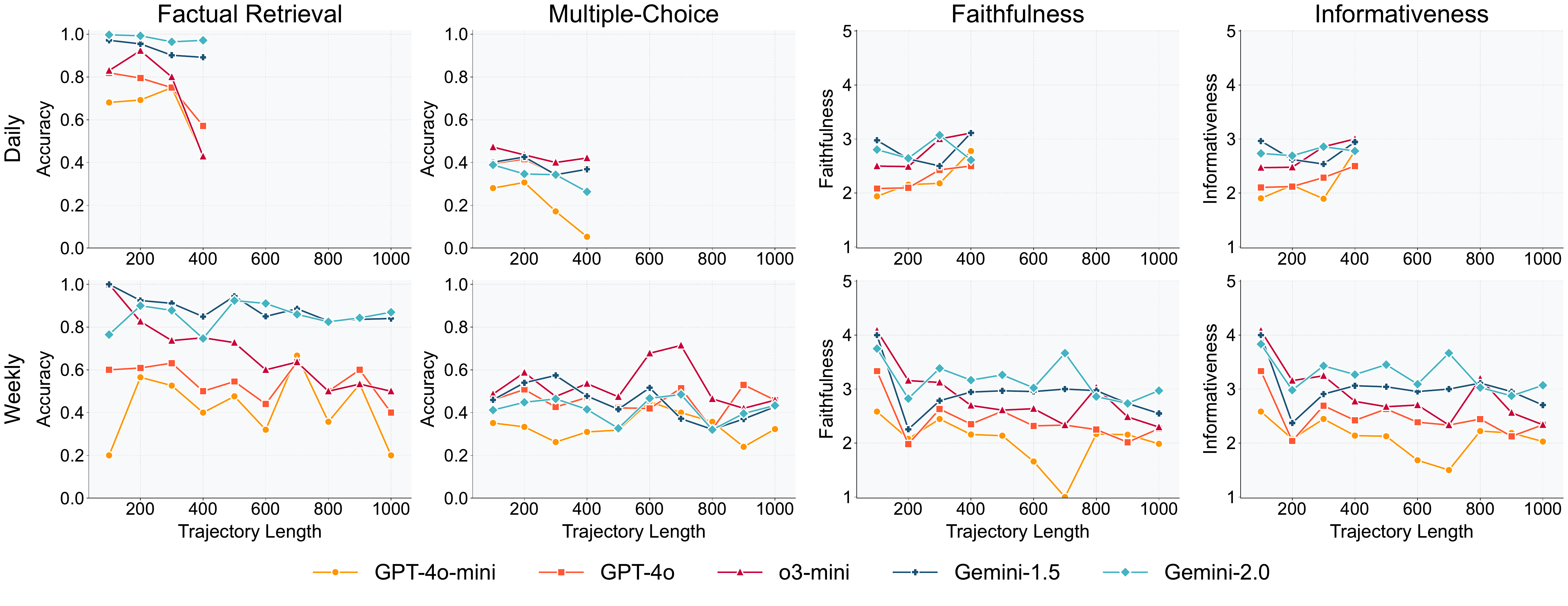}
  \caption{Performance comparisons of closed-source LLMs across varying trajectory lengths}
  \label{fig:accuracy_curve}
\end{figure*}

Fine-tuning demonstrated clear performance improvements over the base models. This was particularly evident in multiple-choice tasks, where the fine-tuned models achieved the highest scores among all models evaluated. While factual retrieval performance also improved, it did not surpass that of the larger base models. On free-form questions, the two series exhibit different trends. The Llama models consistently improved through fine-tuning, whereas the Qwen series shows limited gains, with the base model sometimes outperforming its fine-tuned version on the weekly trajectories. Nevertheless, overall performance on free-form tasks remains limited across all models, highlighting the task's inherent difficulty. This is likely because the task requires high-level reasoning, which may be difficult to acquire just through SFT alone.

\subsection{Impact of Trajectory Lengths}
\label{sec:traj_length_effect}
Previous studies have shown that LLMs often struggle with longer input sequences~\citep{Kim2024-us, Yoon2024-cy}. As shown in Fig.~\ref{fig:accuracy_curve}, we investigated how input trajectory lengths affect mobility question answering performances across different tasks by dividing the daily and weekly trajectory data into bins based on their lengths.

For the factual retrieval questions, GPT models exhibit a sharp drop in accuracy as trajectory length increases. One possible reason for the performance drop is that longer trajectories include more places and events, which increases the number of possible details to choose from. This may make it harder for the models to focus on the most relevant information and retrieve the correct answer. In contrast, Gemini models show a more gradual decline or relatively stable performance across different input lengths.

On the other hand, trajectory length appears to have less impact for the multiple-choice questions. Most models maintain relatively stable accuracy regardless of input length. Interestingly, some models, particularly regarding the weekly trajectories, achieve slightly better accuracy on longer input trajectories, possibly because they at the same time contain richer temporal context. These mixed results suggest that the optimal context length for multiple-choice reasoning may vary depending on the models.

\begin{figure}[h]
  \centering
  \includegraphics[width=\linewidth]{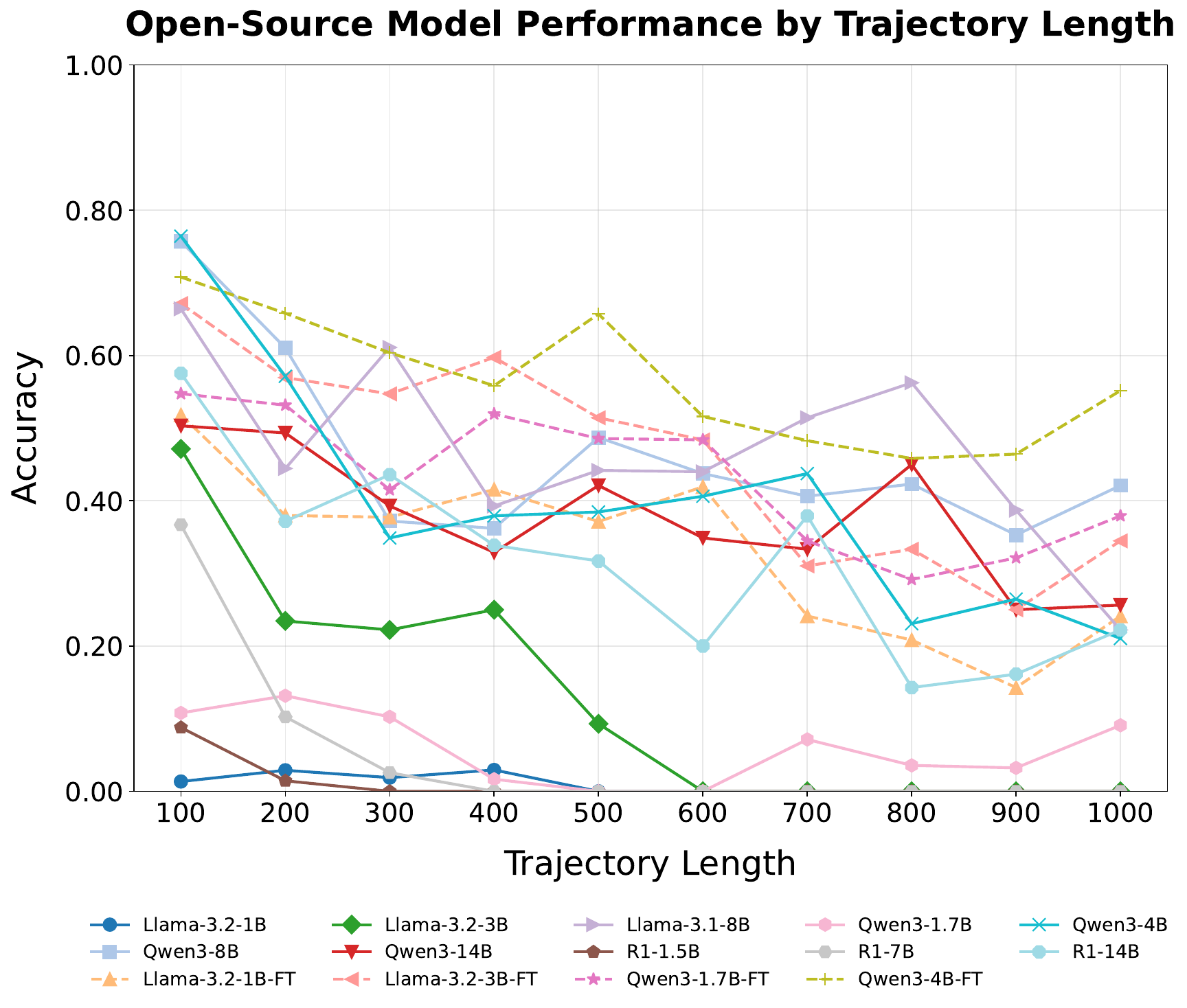}
  \caption{Performance comparisons of open-source LLMs and their fine-tuned versions across varying trajectory lengths}
  \label{fig:open_model_results}
\end{figure}

\balance
\subsection{Effect of Semantic Information}
\label{sec:semantic_info_effect}

For faithfulness and informativeness in free-form questions, the effects of trajectory length differ across conditions. For daily trajectories, scores tend to improve slightly as their length becomes longer, implying that richer input may help models produce more detailed and accurate answers. In contrast, for the weekly trajectories, both metrics degraded as trajectory length increases. Notably, the highest scores (approaching 4) occur on shorter weekly trajectories, where behavior is likely more focused and easier to summarize. Longer sequences may introduce noise or complexity that interferes with coherent generation.

We also performed a trajectory length analysis on the open-source models (Fig.~\ref{fig:open_model_results}), which presents the combined results for factual retrieval and multiple-choice tasks. Similar to the trends observed with the closed models, all ten open-source models showed a consistent degradation in performance as the trajectory length increased. This negative impact was more pronounced in smaller models. Although fine-tuning consistently improved performance over the base models, it failed to resolve the issue of performance degradation with longer sequences. This highlights the inherent difficulty of the underlying problem, which remains a challenge even for specialized models.

Mobility data in real-world applications can sometimes include semantic information such as transportation modes, POIs, or reasons for visits, which may be beneficial for mobility question answering. 
To investigate this, we evaluated multiple-choice accuracy for weekly trajectory data with four input configurations: raw trajectories (None) and trajectories augmented with mobility modes, POIs, or inferred reasons of visit.

\begin{figure}[t]
\centering
\includegraphics[width=\linewidth]{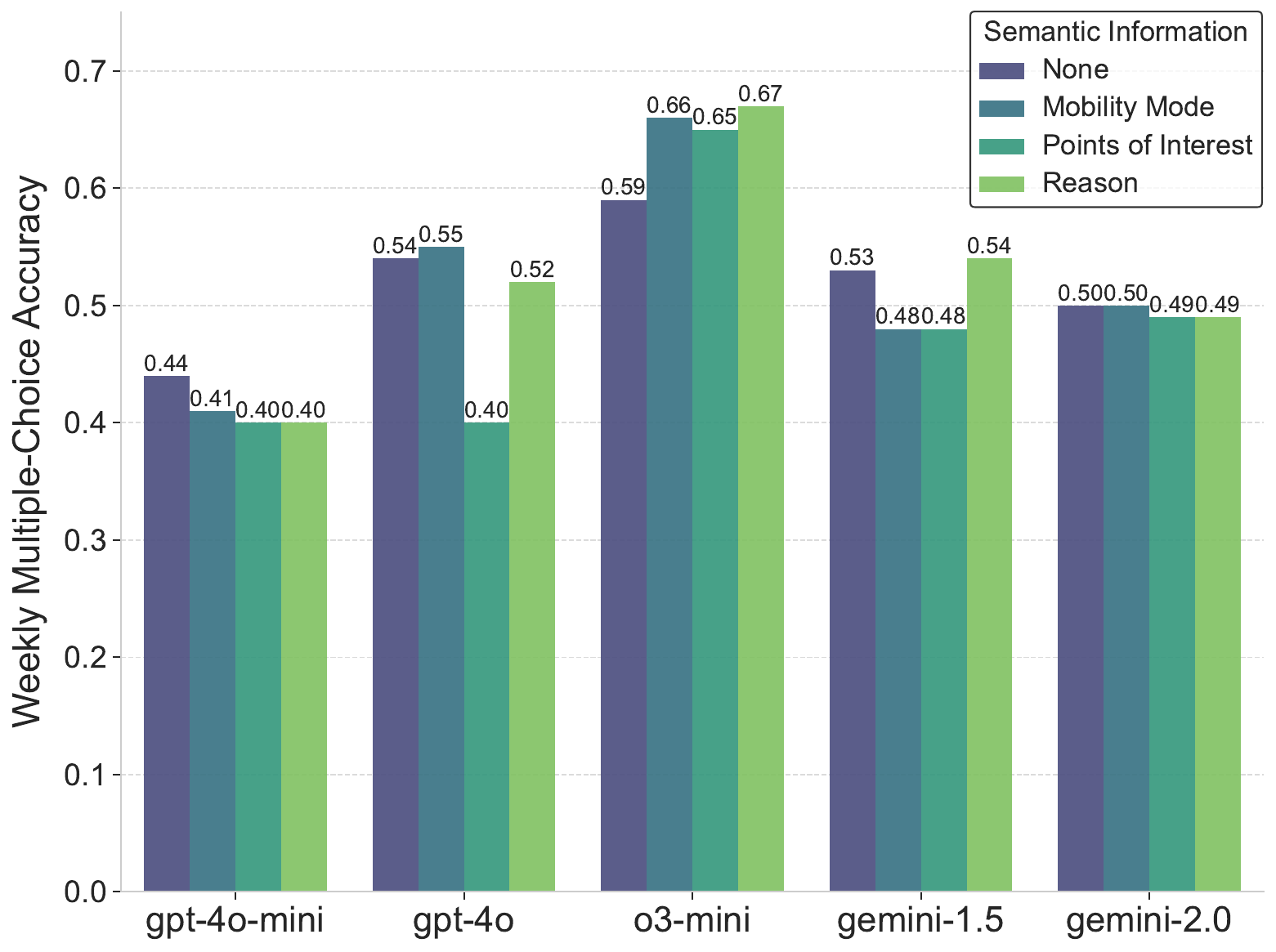}
\caption{Effect of semantic information for multiple-choice question answering performances}
\label{fig:accuracy_semantic}
\end{figure}

As reported in Fig.~\ref{fig:accuracy_semantic}, o3-mini demonstrated substantial improvement with semantic augmentation, achieving its highest accuracy (0.67) when reasons of visit were included. In contrast, gpt-4o-mini performed worse with all types of semantic inputs, potentially due to longer sequences making it difficult for the model to focus on relevant information. Gemini-1.5 and Gemini-2.0 remained relatively stable with minimal gains or losses. While this stability suggests robustness, it also indicates that these models may not fully exploit the added semantic cues under current settings. These results indicate that semantic cues benefit performance only when LLMs can integrate them effectively. Without proper handling, such augmentations may even degrade performance, particularly in smaller models.

\section{Related Work}
\label{sec:related}
The MobQA benchmark dataset intersects with and contributes to several key research areas. We position our work within three complementary domains: (1) mobility data analysis, (2) spatiotemporal reasoning, and (3) LLMs for trajectory and sensor data.

\subsection{Mobility Data Analysis}
Human mobility research has shifted from early statistical and feature engineering-based methods~\citep{Calabrese2010-cm,Gambs2010-ju,Monreale2009-zf,Noulas2012-wl} to deep learning models such as RNNs and Transformers that capture sequential trajectory patterns~\citep{Chung2015-vw,Vaswani2017-nv,Liu2016-bk,Feng2020-rq,Liang2022-sk}. More recently, architectures that embed heterogeneous context, such as urban layout and POIs, have achieved state-of-the-art accuracy on next-location prediction and POI recommendation~\citep{Wang2019-is,Haydari2024-yw,Luca2021-ix}.
Yet existing models, while effective at trajectory prediction, seldom capture the purposes and contextual factors that give human movement its meaning~\citep{Xue2023-dr, Mokbel2024-zr}.

\subsection{Spatiotemporal Reasoning}
Mobility question answering inherently requires sophisticated spatiotemporal reasoning capabilities. Research in temporal reasoning has centered on tasks such as event ordering~\citep{Ning2020-pj, Naik2019-wl}, frequency or duration estimation~\citep{Rajpurkar2016-rp, Zhou2019-tv, Virgo2022-pf}, and temporal causality analysis~\citep{Mirza2014-kn, Zhou2021-up}. Even state-of-the-art LLMs can fail to work on such tasks~\citep{Gupta2023-mn, Yang2023-vy, Jia2024-lt, Chu2024-eb, Wang2024-qq, Xiong2024-mh, Fatemi2025-oy}. In mobility question answering, time appears only as raw timestamps, making the task complex: models must infer temporal landmarks from coordinate streams and relate these signals to broader concepts.

Spatial reasoning is equally demanding. Vision-language models perform well when spatial cues are explicit in images~\citep{Johnson2017-no, Mirzaee2021-jk, Mirzaee2022-ou, Kamath2023-ds, Liu2023-zd, Cheng2024-ue, Cheng2024-le, Shiri2024-df, Lei2025-ry} or video~\citep{Jang2017-bt, Yu2019-aj, Lei2020-dw, Grunde-McLaughlin2021-dy}, yet they still struggle with nuanced relations~\citep{Cheng2024-nb, Fu2025-pb}.
Mobility question answering removes visual context entirely; LLMs must recover spatial relationships directly from sequences of coordinates and timestamps, serving as a substantially harder benchmark than conventional vision tasks~\citep{Majumdar2024-eo}. To succeed, models must reconstruct latent spatial hierarchies, such as neighborhoods, transit corridors, and functional regions, and reason about distance, containment, and orientation without any explicit geographic metadata.

\subsection{Large Language Models for Mobility and Sensor Data}
Applying LLMs to mobility and sensor data is a growing frontier. Recent studies tackle trajectory prediction~\citep{Wang2023-vt, Liang2024-nd, Xu2024-je, Li2024-lc}, synthetic data generation~\citep{Asano2024-dk, Wang2024-re}, and sensor classification~\citep{Endo2023-gx, Li2024-lk, Ji2024-zx, Kim2024-us}. Yet most work remains confined to closed-ended tasks, short sequences (often under 100 events)~\citep{Kim2024-us, Yoon2024-cy}, or enriched inputs, such as check-in POIs~\citep{Wang2023-vt}. Handling long streams of raw coordinates, lacking explicit semantics, continues to challenge current LLMs~\citep{Fang2024-qw, Yoon2024-cy}.

Our work directly addresses these gaps. Mobility question answering introduces a framework for open-ended, natural language question answering based on raw trajectory data composed primarily of coordinates and timestamps, moving beyond simple prediction toward genuine behavioral interpretation.

\section{Conclusion}
\label{sec:conclusion}

We introduced MobQA, the first comprehensive benchmark dataset for evaluating semantic understanding of human mobility through natural language question answering. We formulated the task of mobility question answering, defined complementary types of questions such as factual retrieval, multiple-choice reasoning, and free-form explanation, and carefully constructed the dataset on top of the Geolife GPS dataset. Our extensive experiments reveal the advantages and limitations of various state-of-the-art LLMs in interpreting human mobility data. We believe MobQA will serve as a valuable resource for advancing fundamental research and applications of LLMs in mobility data analysis.

\clearpage

\bibliographystyle{ACM-Reference-Format}
\bibliography{sample-base}

\appendix

\section{Limitations}

\paragraph{Dataset Scope}
The MobQA dataset is built on the Geolife dataset,\footnote{\url{https://www.microsoft.com/en-us/research/publication/geolife-gps-trajectory-dataset-user-guide/}} which provides rich urban trajectory data collected in Beijing. Due to privacy concerns associated with collecting and releasing human mobility data, we chose to use this publicly available dataset that was explicitly consented for research use. While the Geolife dataset offers a valuable starting point, the geographic and demographic scope remains limited. Moreover, our dataset is currently constructed in Japanese. Nevertheless, our data construction framework is inherently language-agnostic and can be extended to other languages and cultural contexts seamlessly without major changes in the annotation and question-answer generation processes. Future work will seek to build multilingual versions of the dataset with more diversity in geography and demography. This will support broader global applicability and enable future studies on how such factors influence mobility understanding.

\paragraph{Evaluation Scope}
This study focuses on three core tasks:factual retrieval, multiple-choice reasoning, and free-form explanation, as a starting point for semantic mobility understanding. While these cover essential capabilities, they do not exhaust the full space of mobility-related reasoning. Future extensions could incorporate dialogue-based interactions, long-term behavior summarization, or subjective tasks such as intent or preference estimation, enabling richer assessment of model understanding.

\section{Reproducibility Measures}
The MobQA dataset is composed of anonymized trajectories, questions, and reference answers, which are derived from the Geolife GPS traces. The Geolife dataset is distributed under the Microsoft Research License Agreement (MSR-LA). We employ the dataset purely for academic and benchmarking purposes, strictly adhering to its terms for non-commercial research use. Crucially, following direct communication with the original Geolife authors, we have obtained explicit written permission to release the generated question-answer pairs, semantic trajectory annotations, and corresponding user IDs. This authorization ensures full compliance with the original dataset's licensing terms while the original GPS traces remain protected under the license agreement.

To ensure full reproducibility under MSR-LA, we adopt an approach where each researcher can individually download and obtain the Geolife license, then reconstruct the complete MobQA dataset using our released components. Specifically, we open-source: (i)~the generated question-answer pairs with corresponding user IDs and trajectory dates, (ii)~the semantic trajectory annotations, (iii)~reconstruction code that generates the complete MobQA dataset from the downloaded Geolife dataset and our released QA pairs and trajectory information, and (iv)~all evaluation pipelines and reproduction scripts, including answer generation and evaluation code (prompts for each task, scoring scripts) required to reproduce our experiments. Note however that differences in LLM API versions may introduce an unavoidable ceiling on exact replicability.

\section{Ethical Considerations}
Human mobility data, while valuable for research, carries inherent potential risks, particularly concerning individual privacy. Misuse of such data could lead to significant privacy infringements. To address these concerns, our study exclusively utilizes \emph{anonymized trajectories} as introduced in Sec.~\ref{subsec:trajectory_data}. This anonymization is a critical measure to ensure that individual identities are protected and that the data cannot be traced back to specific persons, thereby resolving the primary privacy issues associated with using detailed movement data in our experiments.

\section{Justification for Base Dataset Selection}
\label{sec:base_dataset_justification}

Our goal is to enable semantic question answering over trajectories, which requires understanding dense, multi-mode movement nuances. Selecting an appropriate base dataset is therefore critical. We reviewed existing surveys covering publicly available human mobility datasets~\citep{Calabrese2015-io, Zhao2016-ci, Luca2021-ix}. These datasets can be broadly categorized as follows, though most are unsuitable for our goal.

\begin{itemize}[itemsep=1.5pt, topsep=5pt, leftmargin=1.2em, labelsep=0.6em, after=\vspace{0.35\baselineskip}]
\item\textbf{Single-Mode Datasets} (\eg, NYC Yellow Taxi~\citep{NYC2021}, T-Drive~\citep{Yuan2010-lb}) are limited to a single mode of transport, primarily taxis. This makes them unsuitable for analyzing multi-modal mobility patterns or personal routine behaviors.
\item\textbf{Sparse Check-in Logs} (\eg, Foursquare~\citep{Yang2015-wj, Yang2016-np}, Gowalla~\citep{Cho2011-dz}) consist of check-in logs recorded only upon arrival at specific locations. This creates extremely sparse data points. While useful for tasks like point-of-interest recommendation, they lack the continuous path information necessary for understanding the \textit{how} and \textit{why} of movement between locations.

\item\textbf{Coarse-Grained GPS Logs} intentionally reduce spatiotemporal resolution, often for privacy. Trajectories are discretized into large grid cells (\eg, 500m x 500m) and long time intervals (\eg, 30 minutes), as seen in YJMob100K~\citep{Yabe2024-ui}. This granularity differs fundamentally from the dense trajectories our work requires, as fine-grained details like short stops, mode changes, or precise routes cannot be inferred from such coarse data.

\item\textbf{Infrastructure-based Localization Traces} (\eg, MIT Reality Mining~\citep{Eagle2006-aj}, Orange D4D~\citep{Blondel2012-ra}) rely on fixed infrastructure (e.g., Wi-Fi access points or cell towers), resulting in coarse spatial granularity and sparse temporal resolution. Thus, they lack the fine-grained trajectories required for detailed sequential mobility analysis.

\end{itemize}
Indeed, our preliminary experiments with sparse data, including Foursquare check-ins and YJMob100K's coarse-grained GPS, confirmed these limitations. We found that from these abstract trajectories, even human experts were unable to reliably infer the semantic nuances required for our QA task. While such datasets may be sufficient for trajectory prediction or generation, they cannot support the deep, context-aware understanding that our natural language-based benchmark is designed to evaluate.

This systematic elimination identified the \emph{Geolife GPS dataset}~\citep{Zheng2008-gf,Zheng2010-lk,Zheng2011-mg} as the only publicly available option that meets our stringent requirements. Geolife is the ideal foundation for our MobQA dataset for the following reasons:
\begin{itemize}[itemsep=1.5pt, topsep=5pt, leftmargin=1.2em, labelsep=0.6em, after=\vspace{0.35\baselineskip}]
  \item It provides high spatiotemporal resolution, enabling detailed trajectory analysis.
  \item It includes verified labels for 11 transport modes (\eg, walking, biking, subway), covers diverse behaviors (\eg, commutes, leisure), and spans over five years for 182 users.
  \item It is openly accessible and comprehensively documented, ensuring full replicability.
\end{itemize}

\section{Evaluation Methodology Details}
\label{sec:evaluation_methodology_details}

\subsection{Evaluation Metric for Free-Form Tasks}
\label{subsec:evaluation_methodology_comparison}

To select an appropriate evaluation metric for free-form questions, we compared conventional automatic metrics with the recent LLM-as-a-judge framework. Specifically, we benchmarked following widely used metrics against human judgments:

\begin{itemize}[itemsep=1.5pt, topsep=5pt, leftmargin=1.2em, labelsep=0.6em, after=\vspace{0.35\baselineskip}]
  \item \textbf{BLEU}~\citep{Papineni2002-cr}: A precision-oriented metric that measures $n$-gram overlap between generated and reference texts, widely adopted in machine translation evaluation.
  \item \textbf{ROUGE}~\citep{lin-2004-rouge}: A recall-focused metric that computes $n$-gram overlap (\eg, ROUGE-2) or the longest common subsequence (ROUGE-L), commonly used for summarization tasks.
  \item \textbf{BERTScore}~\citep{Zhang2020-xb}: A semantic-similarity metric that leverages contextualized BERT embeddings to capture meaning beyond surface-level token matches.
  \end{itemize}
Tab.~\ref{tab:correlation_all} presents the results from benchmarking these metrics on both daily and weekly subsets of our data. Our findings indicate that traditional automatic metrics correlate poorly with human judgments. ROUGE-2 and ROUGE-L show particularly weak or even negative correlations, while BLEU and BERTScore achieve modest positive correlations that are nevertheless substantially lower than those of LLM-based evaluators.

In contrast, LLM-as-a-judge evaluators show substantially stronger agreement with human assessments. GPT-4o-mini achieves the highest faithfulness correlation on daily data, while Gemini models excel on weekly data. Gemini 2.0 yields the highest informativeness correlations, exceeding 0.57 for both temporal scales. Overall, LLM-based judges achieve correlations in the 0.40-0.60 range, substantially outperforming automatic metrics. Given the inadequacy of conventional metrics in this context, we adopt LLM-as-a-judge assessments for faithfulness and informativeness throughout our study. This approach aligns with recent best practices for evaluating open-ended generation tasks~\citep{Zheng2023-au, Gu2024-kt}.

\begin{table}[t]
  \centering
  \caption{Comparing Automatic Metrics and LLM-as-a-Judge: Correlation Coefficients for Faithfulness and Informativeness}
  \label{tab:correlation_all}
  \footnotesize
  {\setlength{\tabcolsep}{4pt}%
  \begin{tabular}{lcccc}
    \toprule
    \multirow{2}{*}{Eval Method} & \multicolumn{2}{c}{Faithfulness} & \multicolumn{2}{c}{Informativeness} \\
    \cmidrule(lr){2-3} \cmidrule(lr){4-5}
    & One Day & Weekly & One Day & Weekly \\
    \midrule
    \multicolumn{5}{c}{\textit{Automatic Metrics}} \\
    \midrule
    BLEU                  & 0.081 / 0.206 & 0.185 / 0.230 & 0.076 / 0.175 & 0.113 / 0.154 \\
    ROUGE-2               & -0.191 / -0.093 & -0.024 / 0.001 & -0.272 / -0.206 & -0.017 / 0.010 \\
    ROUGE-L               & -0.003 / -0.006 & 0.110 / 0.136 & -0.122 / -0.151 & 0.167 / 0.183 \\
    BERTScore             & 0.239 / 0.252 & 0.086 / 0.114 & 0.279 / 0.294 & 0.136 / 0.151 \\
    \midrule
    \multicolumn{5}{c}{\textit{LLM-as-a-Judge}} \\
    \midrule
    GPT-4o-mini           & 0.502 / \textbf{0.562} & 0.278 / 0.349 & 0.554 / 0.605 & 0.402 / 0.425 \\
    GPT-4o                & \textbf{0.507} / 0.524 & 0.401 / 0.492 & 0.521 / 0.511 & 0.513 / 0.537 \\
    Gemini 1.5            & 0.370 / 0.378 & \textbf{0.513} / \textbf{0.548} & 0.544 / 0.593 & 0.502 / 0.521 \\
    Gemini 2.0            & 0.368 / 0.401 & 0.494 / 0.535 & \textbf{0.584} / \textbf{0.608} & \textbf{0.577} / \textbf{0.596} \\
    \bottomrule
  \end{tabular}
  }%
\end{table}

\subsection{Self-Enhancement Bias in LLM-as-a-Judge}

Recent studies have highlighted self-enhancement bias in LLM-as-a-judge evaluations, where evaluators may favor responses from their own model family~\citep{Zheng2023-au, Dai2024-cr, Ye2025-yh}. To mitigate this bias and ensure the validity of 
our results,, we conducted a comparative analysis using both our primary evaluator (Gemini-2.0-Flash) and GPT-4o as a secondary evaluator from a different model family.

\begin{table}[t]
  \centering
  \caption{Performance of LLMs on Free-form Questions (Faithfulness \& Informativeness). Results are shown as Gemini-2.0-Flash/GPT-4o evaluation scores. \textbf{\underline{bold + Underline}} indicates the best score, \textbf{Bold} indicates the second best, and \underline{Underline} indicates the third best.}
  \label{tab:eval_comparison}
  \begin{tabular}{lcccc}
    \toprule
    \multirow{2}{*}{Model} & \multicolumn{2}{c}{Faithfulness} & \multicolumn{2}{c}{Informativeness} \\
    \cmidrule(lr){2-3}\cmidrule(lr){4-5}
    & One Day & Weekly & One Day & Weekly \\
    \midrule
    GPT-4o-mini        & 2.01/2.07 & 1.98/1.92 & 1.97/1.82 & 2.01/1.83 \\
    GPT-4o             & 2.15/2.16 & 2.27/2.25 & 2.15/2.07 & 2.35/2.23 \\
    o3-mini            & \underline{2.46}/\underline{2.42} & \underline{2.74}/\textbf{2.70} & \underline{2.43}/\underline{2.08} & \underline{2.81}/\underline{2.46} \\
    Gemini 1.5         & \textbf{\underline{3.12}}/\textbf{2.94} & \textbf{2.81}/\underline{2.58} & \textbf{\underline{3.08}}/\textbf{2.42} & \textbf{2.91}/\textbf{2.38} \\
    Gemini 2.0         & \textbf{2.96}/\textbf{2.79} & \textbf{\underline{3.09}}/\textbf{2.69} & \textbf{2.91}/\textbf{2.30} & \textbf{\underline{3.22}}/\textbf{\underline{2.48}} \\
    \bottomrule
  \end{tabular}
\end{table}

Our analysis revealed a consistency in the assessment patterns of both 
evaluators. As shown in Tab.~\ref{tab:eval_comparison}, the performance ranking remained stable for both faithfulness and informativeness metrics, regardless of whether Gemini-2.0-Flash or GPT-4o served as judge. Notably, the Gemini family consistently received higher scores than other models across 
the board.

To statistically validate this inter-evaluator agreement, we analyzed the correlation between their scores. The results confirmed a very strong positive correlation, with a Pearson correlation coefficient of $r=0.837$ and a Spearman's rank correlation of $\rho=0.845$. This high correlation provides compelling evidence that our evaluation framework is robust and that the observed performance differences reflect genuine model capabilities rather than artifacts of self-enhancement bias.

\section{Semantic Annotation Details}
\label{sec:semantic_traj}
\begin{figure*}[t]
    \centering
    \includegraphics[width=\linewidth]{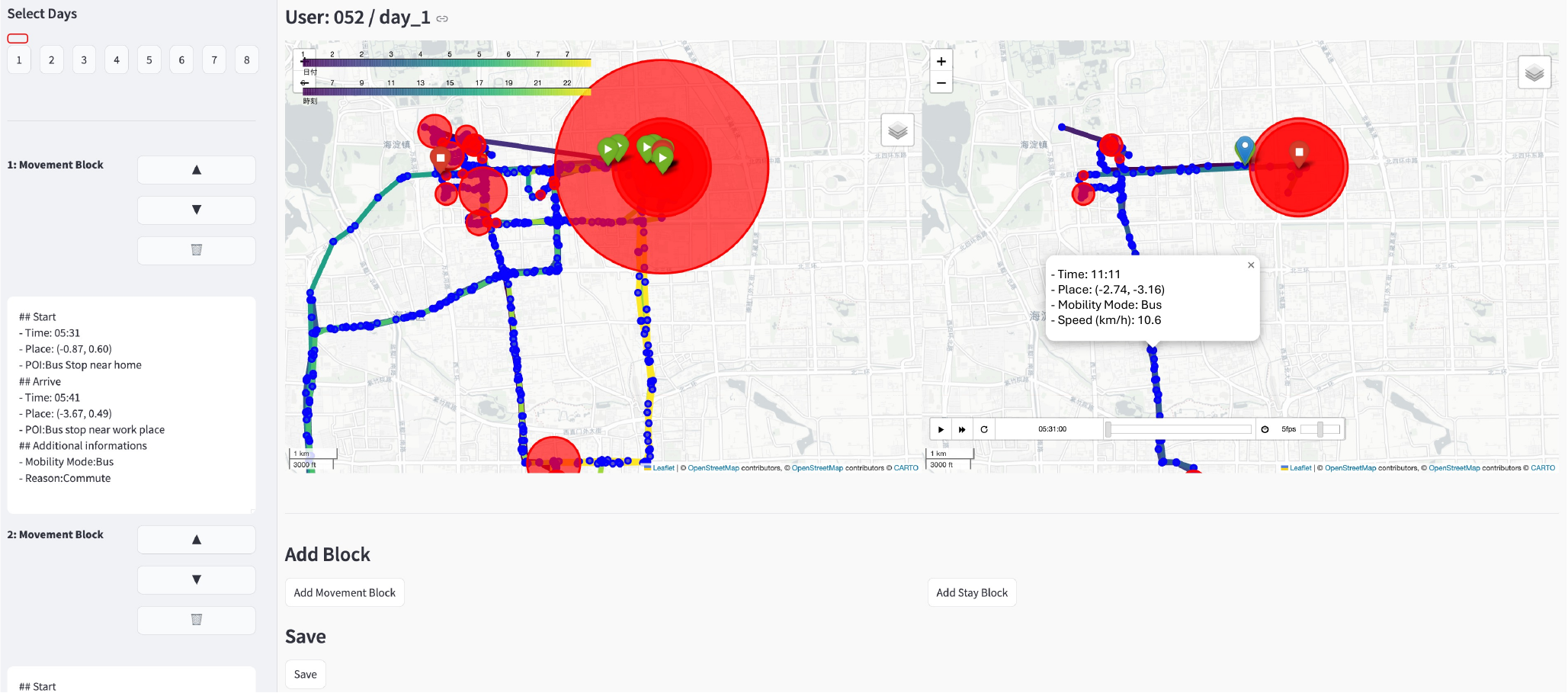}
    \caption{Screenshot of the Streamlit-based semantic trajectory annotation interface. The leftmost pane lets annotators switch among eight consecutive days, providing constant week-level context. In the main map view, automatically detected stay points appear as red circles whose radii reflect duration, whereas individual GPS fixes are shown as blue dots; inter-dot spacing intuitively conveys movement speed. Hover tooltips reveal timestamp, coordinates, speed, and user-reported mobility mode, and playback controls enable frame-by-frame inspection of the trajectory. Below the map, editable movement and stay blocks can be inserted, reordered, or deleted, streamlining fine-grained semantic refinement.}
    \label{fig:annotation_tool} %
\end{figure*}
\label{app:annotation}

This section provides detailed information about the semantic annotation process described in Sec.~\ref{subsec:semantic_annotation}. Three expert annotators specialized in mobility-data annotation processed a total of 136 trajectories (17 users $\times$ 8 consecutive days), spending approximately 60 hours in total (excluding training and instruction), averaging around 26 minutes per day-trajectory.

\subsection{Annotation Tool}
\label{subsec:annotation_tool}
The custom annotation tool (Fig.~\ref{fig:annotation_tool}) was developed based on Streamlit\footnote {https://streamlit.io/}, providing an interactive environment to facilitate precise semantic labeling.
The interface consists of two main components: a contextual view of eight consecutive days displayed on the left pane, and a detailed visualization of the selected day's trajectory on the right pane. The interactive trajectory visualization was built using MovingPandas\footnote{\url{https://movingpandas.org/}}, allowing annotators to intuitively interpret movement characteristics.
Specifically, rule-based automatically identified stationary points are depicted by red circles, whose sizes indicate the duration of stays.\footnote{These stay durations were carefully redefined by annotators during the annotation process to ensure accuracy.}
Blue dots represent individual GPS fixes, with inter-dot spacing visually encoding travel speed—dense clusters signify slower movements or stops, whereas greater spacing indicates faster travel segments.
The interface also allows annotators to zoom in and out of areas of interest, enabling detailed examination of specific locations and movements while maintaining the broader context of the trajectory.
Additionally, annotators can access precise information, such as timestamps, geographic coordinates, speed, and user-reported mobility modes through hover tooltips. Playback functionalities, such as frame-by-frame navigation, further support accurate segmentation.

\subsection{Annotator Details}
\label{subsec:annotator}
The annotation team consisted of three professional annotators (two female, one male) employed within the authors' organization.
All annotators were thoroughly briefed on the purpose of the study, including how the annotated data would be used and its intended public release. They provided informed consent for their annotations to be included in the publicly available dataset.
While we determined that the annotation process posed minimal risks to the annotators (primarily limited to standard ergonomic considerations associated with computer work), we ensured transparent communication about all aspects of the project. The annotators acknowledged their understanding of both the project goals and the public nature of their contributions to the mobility dataset.

\subsection{Annotation Process}
The annotation protocol involved two structured training phases. Initially, annotators participated in a one-hour session, including detailed instructions and hands-on practice guided individually by the lead author using pre-annotated exemplar trajectories. Feedback on tool usability and instruction clarity was collected after this phase. Based on this input, guidelines were clarified, and the annotation tool interface was refined accordingly. Subsequently, annotators underwent a second one-hour instructional session before proceeding with the main annotation tasks.

Quality assurance was conducted rigorously, with the lead author performing a final validation of all semantic trajectories. This review ensured consistency in mobility modes, segment boundaries aligned precisely to minute-level resolution, and completeness by identifying and correcting potential omissions such as brief transitions between mobility modes. This meticulous verification step guaranteed high-fidelity semantic trajectories suitable for reliable downstream applications in mobility-related question answering.

\section{Manual Answer Filtering}
\label{sec:manual_answer_filtering}
To ensure the quality of ground-truth answers to free-form questions, we employ a manual filtering process, which verifies whether the generated answers are faithful to the original trajectory data and contain information beneficial to the user.

\subsection{Annotators}
Annotators for the answer filtering task were recruited through a crowdsourcing platform named Lancers.\footnote{\url{https://www.lancers.jp}} Initially, we hired nine annotators via an open call and conducted a preliminary annotation task. We then selected the three annotators who demonstrated exceptionally high annotation quality to undertake the main annotation work in this filtering phase. Every annotator was explained on the study purpose and data usage, acknowledged risks, and consented to their annotation being included in the public dataset.

\subsection{Annotation Tool}

We extended the annotation tool developed in Sec.~\ref{subsec:annotation_tool} to visualize additional information necessary for the answer filtering process, such as stop points, POIs, mobility modes, as well as questions and generated answers.
This customized interface enabled annotators to efficiently assess answer quality by simultaneously referencing the original trajectory data along with its semantic context.

\subsection{Evaluation Process and Metrics}
The quality of each answer for the free form questions was evaluated based on faithfulness and informativeness scores defined in Sec.~\ref{subsec:evaluation_methodology}.
Prior to the annotation task, annotators were informed the evaluation protocol with a document that explained the use of the annotation tool, detailed the evaluation metrics, and provided three illustrative examples for different scores of faithfulness and three for informativeness to clarify the scoring criteria. Furthermore, annotators were encouraged to maintain open communication to address any questions or ambiguities they encountered during the process.

Each response was rated on a 1-5 scale for both faithfulness and informativeness. To be finally included in the dataset, an answer had to achieve a score of 4 or higher on both metrics. Annotators first underwent a calibration phase using gold-standard examples scored by experts to ensure a unified understanding of the evaluation criteria. The actual evaluation was conducted using the aforementioned custom interface, where annotators could simultaneously view the raw trajectory data, the semantic trajectory, the question, and the model-generated answer while scoring. As a result of this filtering process, approximately 90\% (around 1,800) of the initially generated 2,000 answers met the acceptance criteria. This high acceptance rate is likely attributed to the benefit of providing semantic information alongside raw trajectory data to the LLM (GPT-4o) during answer generation.

\section{Experiment Details}
\label{sec:experiment_detail}

\subsection{Model Details}
We benchmarked five recent closed-source large language models (LLMs): gpt-4o-mini-2024-07-18\footnote{https://platform.openai.com/docs/models/gpt-4o-mini}, gpt-4o-2024-11-20\footnote{https://platform.openai.com/docs/models/gpt-4o}, and o3-mini-2025-01-31\footnote{https://openai.com/index/openai-o3-mini/} through Azure OpenAI Service\footnote{https://learn.microsoft.com/en-us/azure/ai-services/openai/azure-government}, and gemini-1.5-flash-002\footnote{https://cloud.google.com/vertex-ai/generative-ai/docs/models/gemini/1-5-flash} and gemini-2.0-flash-001\footnote{https://cloud.google.com/vertex-ai/generative-ai/docs/models/gemini/2-0-flash} via Vertex AI Platform\footnote{https://cloud.google.com/vertex-ai}. We also evaluate ten open models to provide a comprehensive comparison: Llama-3.2-1B-Instruct\footnote{\url{https://huggingface.co/meta-llama/Llama-3.2-1B-Instruct}}, Llama-3.2-3B-Instruct\footnote{\url{https://huggingface.co/meta-llama/Llama-3.2-3B-Instruct}}, Llama-3.1-8B-Instruct\footnote{\url{https://huggingface.co/meta-llama/Llama-3.1-8B-Instruct}}, Qwen3-1.7B\footnote{\url{https://huggingface.co/Qwen/Qwen3-1.7B}}, Qwen3-4B\footnote{\url{https://huggingface.co/Qwen/Qwen3-4B}}, Qwen3-8B\footnote{\url{https://huggingface.co/Qwen/Qwen3-8B}}, Qwen3-14B\footnote{\url{https://huggingface.co/Qwen/Qwen3-14B}},  DeepSeek-R1-Distill-Qwen-1.5B\footnote{\url{https://huggingface.co/deepseek-ai/DeepSeek-R1-Distill-Qwen-1.5B}}, DeepSeek-R1-Distill-Qwen-7B\footnote{\url{https://huggingface.co/deepseek-ai/DeepSeek-R1-Distill-Qwen-7B}}, and DeepSeek-R1-Distill-Qwen-14B\footnote{\url{https://huggingface.co/deepseek-ai/DeepSeek-R1-Distill-Qwen-14B}}. All open models were loaded from their official Hugging Face repositories and evaluated using a single A100 GPU configuration. For all experiments, we set the temperature parameter to 0.7 to balance predictability with response diversity. The exact prompts used for factual retrieval, multiple-choice questions, free-form questions, and the LLM-as-a-judge evaluation are detailed in Fig.~\ref{fig:factual_retrieval_prompt}, Fig.~\ref{fig:multiple_choice_prompt}, Fig.~\ref{fig:free_form_prompt}, and Fig.~\ref{fig:llm_judge_prompt}, respectively.

\begin{figure*}[th]
  \centering
  \begin{tcolorbox}[title=Prompt for Factual Retrieval Questions, width=1\textwidth, halign=center]
  \footnotesize
      \begin{lstlisting}[breaklines=true,basicstyle=\small\ttfamily, columns=flexible]
You are provided with anonymized trajectory data. Your task is to extract all explicit and verifiable facts exclusively from the trajectory data. Focus on a precise and comprehensive extraction without any interpretation.

Data Overview:
- Trajectory Data: Contains positions given in kilometers along with other explicit details.
- Privacy Note: The data has been normalized to protect personal details.

Instructions:
1. Accurately list every explicit fact present in the trajectory data, including numerical values and units.
2. Present your findings as a bullet list under the header "extracted fact:".
3. Avoid any added commentary, inference, or interpretation.

Data Provided:
# Trajectory Data
{trajectory}

# Question
{question}

Your output must be entirely data-driven.
      \end{lstlisting}
  \end{tcolorbox}
  \caption{Prompt template used for factual retrieval tasks in our experiments. The template includes instructions for the model to extract precise information from trajectory data, with placeholders for the actual trajectory data and question.}
  \label{fig:factual_retrieval_prompt}
\end{figure*}

\begin{figure*}[th]
  \centering
  \begin{tcolorbox}[title=Prompt for Multiple-Choice Questions, width=1\textwidth, halign=center]
  \footnotesize
      \begin{lstlisting}[breaklines=true,basicstyle=\small\ttfamily, columns=flexible]
You are provided with anonymized trajectory data. Your task is to analyze the provided information and select the most appropriate answer option for the multiple-choice question below. Your analysis must be fully data-driven and supported by clear, logical reasoning, ensuring that every factual detail is extracted solely from the trajectory data.

**Data Overview:**
- **Trajectory Data:** Positions are given in kilometers.
- **Privacy Note:** The data has been normalized to protect personal details.

**Your Objective:**
- Use the trajectory data to construct a comprehensive answer by selecting the correct option for the multiple-choice question below. All extracted information and conclusions must come exclusively from the trajectory data.

**Instructions for Analysis and Response:**
1. **Extract Key Facts:** Identify and list all objective details extracted solely from the trajectory data.
2. **Develop Your Detailed Reasoning:** Provide a detailed explanation (reason_with_annotation_and_feedback) based solely on the trajectory data.
3. **Develop Your Plain Reasoning:** Provide a detailed explanation (reason) based solely on the trajectory data.
4. **Provide a Final Answer:** Select the option that best answers the multiple-choice question directly, basing your choice exclusively on the trajectory data.
5. **Output Format:** Your final response must include the following items:
 - **extracted fact:** Objective details drawn exclusively from the trajectory data.
 - **reason_with_annotation_and_feedback:** Detailed reasoning based solely on the trajectory data.
 - **reason:** Detailed reasoning based solely on the trajectory data.
 - **answer:** Your final, selected answer option.

**Additional Guidelines:**
- Use precise, logical, and unambiguous language.
- Incorporate common sense reasoning.
- Do not introduce any assumptions or external information.

**Data Provided:**
# Trajectory Data
{trajectory}

# Question
{question}

# Choices
{choices}   

Now, please analyze the data and select the appropriate option for the multiple-choice question.
      \end{lstlisting}
  \end{tcolorbox}
  \caption{Prompt template used for multiple-choice questions in our experiments. The template includes instructions for the model to analyze trajectory data and select the most appropriate answer from given options, with placeholders for the actual trajectory data, question, and answer choices.}
  \label{fig:multiple_choice_prompt}
\end{figure*}

\begin{figure*}[th]
  \centering
  \begin{tcolorbox}[title=Prompt for Free-Form Questions, width=1\textwidth, halign=center]
  \footnotesize
      \begin{lstlisting}[breaklines=true,basicstyle=\small\ttfamily, columns=flexible]
You are provided with anonymized trajectory data. Your task is to analyze the provided information and answer the free-form question below. Your analysis must be fully data-driven and supported by clear, logical reasoning, while ensuring that every factual detail is extracted solely from the trajectory data.

**Data Overview:**
- **Trajectory Data:** Positions are given in kilometers.
- **Privacy Note:** The data has been normalized to protect personal details.

**Your Objective:**
- Use the trajectory data to construct a comprehensive answer to the question below. All extracted information and conclusions must come exclusively from the trajectory data.

**Instructions for Analysis and Response:**
1. **Extract Key Facts:** Identify and list all objective details extracted solely from the trajectory data.
2. **Develop Your Detailed Reasoning:** Provide a detailed explanation in Japanese (reason_with_annotation_and_feedback) based solely on the trajectory data.
3. **Develop Your Plain Reasoning:** Provide a detailed explanation in Japanese (reason) based solely on the trajectory data.
4. **Provide a Final Answer:** Answer the question directly in Japanese, basing your answer exclusively on the trajectory data.
5. **Output Format:** Your final response must include the following items:
 - **extracted fact:** Objective details drawn exclusively from the trajectory data.
 - **reason_with_annotation_and_feedback:** Detailed reasoning based solely on the trajectory data.
 - **reason:** Detailed reasoning based solely on the trajectory data.
 - **answer:** Your final, direct answer to the question in Japanese.

**Additional Guidelines:**
- Use precise, logical, and unambiguous language.
- Incorporate common sense reasoning.
- Do not introduce any assumptions or external information.

**Data Provided:**
# Trajectory Data
{trajectory}

# Question
{question}

**IMPORTANT:** Your final answer and reason must be written in Japanese.

Now, please analyze the data and answer the question.
      \end{lstlisting}
  \end{tcolorbox}
  \caption{Prompt template used for free-form questions in our experiments. The template includes instructions for the model to analyze trajectory data and provide comprehensive answers to open-ended questions, with placeholders for the actual trajectory data and question. The model is specifically instructed to provide its final answer and reasoning in Japanese.}
  \label{fig:free_form_prompt}
\end{figure*}

\begin{figure*}[th]
  \centering
  \begin{tcolorbox}[title=Prompt for LLM-as-a-Judge Evaluation, width=1\textwidth, halign=center]
  \scriptsize
      \begin{lstlisting}[breaklines=true,basicstyle=\tiny\ttfamily, columns=flexible]
You are provided with anonymized trajectory data, an annotation, a question, a reference answer, and a candidate answer. Your task is to critically evaluate the candidate answer by comparing it with the reference answer and verifying its consistency with both the trajectory data and the annotation. Employ an extremely strict and conservative grading approach:
- Assign a score of 4 or 5 only when the candidate answer flawlessly infers the poi, reason, and mobility mode, and every provided fact exactly matches the trajectory data and annotation.
- Assign a score of 3 if the candidate answer is partially correct or has minor, non-critical issues, meaning it is neither flawlessly correct nor severely flawed.
- Assign a score of 1 or 2 if there are significant discrepancies, or any incorrect inference.

Evaluation Criteria:
1. Faithfulness:
- Rigorously verify that the candidate answer reproduces every factual detail (such as locations, timestamps, and movement patterns) exactly as presented in the trajectory data and supported by the annotation.
- Confirm that the inferences regarding poi, reason, and mobility mode are entirely correct.
- **Assign a high score (4 or 5) if the candidate answer is fully faithful to the trajectory data and annotation without any error.**
- **Assign a score of 3 if there are minor inaccuracies or omissions that do not fundamentally alter the overall factual representation, or if the answer is mostly correct but not flawless.**
- **Assign a low score (1 or 2) if significant inaccuracies, misrepresentations, or omissions are identified, or if any critical inference regarding poi, reason, or mobility mode is incorrect.**

2. Informativeness:
- Critically assess whether the candidate answer provides a comprehensive and well-reasoned explanation that is on par with the reference answer, especially regarding inferences on poi, reason, and mobility mode.
- **Assign a high score (4 or 5) if and only if the reasoning is impeccable and all facts are completely consistent with the provided data, and the candidate answer is appropriate to the question and is supported by rich details on poi, reason, and mobility mode.**
- **Assign a score of 3 if the reasoning is somewhat superficial or lacks depth compared to the reference, but is not misleading or incorrect, or if it provides some relevant details but is not comprehensive.**
- **Assign a low score (1 or 2) if the answer is superficial, lacks analytical depth, or fails to properly integrate and critique the trajectory data with the provided annotations, or if the reasoning is incorrect or misleading.**

***STRONGLY MANDATE: IF THE REFERENCE ANSWER AND THE CANDIDATE ANSWER EXPRESS SIMILAR THOUGHTS, YOU MUST ASSIGN HIGH SCORES (4 OR 5) TO BOTH FAITHFULNESS AND INFORMATIVENESS WITHOUT EXCEPTION.***

**IMPORTANT NOTE**: Faithfulness and informativeness are **COMPLETELY SEPARATE** evaluation dimensions. It is **ABSOLUTELY ACCEPTABLE** to assign **VASTLY DIFFERENT** scores to these criteria (e.g., **VERY HIGH** faithfulness but **VERY LOW** informativeness, or vice versa) if that accurately reflects the quality of the candidate answer. **DO NOT** feel obligated to assign similar scores to both dimensions.

Output Format:
- extracted_fact: Key factual elements extracted from the trajectory data.
- reason: Detailed reasoning behind your evaluation.
- faithfulness: A score from 1 to 5.
- informativeness: A score from 1 to 5.

Data Provided:
# Trajectory Data
{trajectory}

# Question
{question}

# Reference Answer
{reference_answer}

# Candidate Answer
{answer}

Now, please evaluate the candidate answer.
      \end{lstlisting}
  \end{tcolorbox}
  \caption{Prompt template used for LLM-as-a-judge evaluation in our experiments. The template includes instructions for the model to critically evaluate candidate answers by comparing them with reference answers and verifying consistency with trajectory data and semantic information.}
  \label{fig:llm_judge_prompt}
\end{figure*}

\subsection{Supervised Fine-tuning Details}

To compare the performance of specialized models against general-purpose ones, we conducted Supervised Fine-tuning (SFT)~\citep{Ouyang2022-gr, Wang2023-st}. This process finetuned the models in a format where the model learn to generate the answer for a MobQA task from a given trajectory and question. We applied SFT to four selected open-source models: Llama-3.2-1B-Instruct, Llama-3.2-3B-Instruct, Qwen3-1.7B, and Qwen3-4B. For efficient fine-tuning, we employed Quantized Low-Rank Adaptation (QLoRA)~\citep{Dettmers2023-vs}, a technique that significantly reduces memory usage by quantizing the model and training only a small number of adapter weights.
The models were subsequently fine-tuned on the MobQA training split for three epochs with a batch size of 2. The complete set of hyperparameters is detailed in Table~\ref{tab:sft_hyperparams}.
The fine-tuning was implemented using the SFTTrainer from the TRL library (\url{https://huggingface.co/docs/trl/en/sft_trainer}). For any hyperparameters not specified in the table, the default values from the SFTTrainer were used.

\begin{table}[t]
  \centering
  \caption{Training hyperparameters for Supervised Fine-tuning (SFT) experiments.}
  \label{tab:sft_hyperparams}
  \begin{tabular}{lc}
  \toprule
  \textbf{Parameter} & \textbf{Value} \\
  \midrule
  Training batch size & 2 \\
  Number of epochs & 3 \\
  Learning rate & 2e-4 \\
  Weight decay & 0.01 \\
  Learning rate scheduler & Linear \\
  LoRA rank & 32 \\
  LoRA alpha & 32 \\
  LoRA dropout & 0.1 \\
  4-bit quantization & True \\
  \bottomrule
  \end{tabular}
\end{table}

\section{Full Dataset Result}
\label{sec:full_dataset_result}
Tab.~\ref{tab:full_dataset_result} shows the evaluation results of closed-source LLMs for all the test data including longer trajectories that exceeds our GPU memory limit for open-source models.

\begin{table}[h!]
  \centering
  \caption{Performance comparisons of closed-source LLMs}
  \label{tab:full_dataset_result}
  \scriptsize
  {\setlength{\tabcolsep}{3.1pt}%
  \begin{tabular}{lcccccccc}
    \toprule
    \multirow{2}{*}{Model} & \multicolumn{2}{c}{Factual Retrieval} & \multicolumn{2}{c}{Multiple-Choice} & \multicolumn{2}{c}{Faithfulness} & \multicolumn{2}{c}{Informativeness} \\
    \cmidrule(lr){2-3} \cmidrule(lr){4-5} \cmidrule(lr){6-7} \cmidrule(lr){8-9}
           & Daily & Weekly & Daily & Weekly & Daily & Weekly & Daily & Weekly \\
    \midrule
    GPT-4o-mini & 0.690 & 0.427 & 0.294 & 0.352 & 2.01 & 1.98 & 1.97 & 2.01 \\
    GPT-4o      & 0.812 & 0.580 & 0.413 & \textbf{0.476} & 2.15 & 2.27 & 2.15 & 2.35 \\
    o3-mini     & 0.824 & 0.644 & 0.472 & 0.473 & 2.46 & 2.74 & 2.43 & 2.81 \\
    Gemini 1.5  & 0.963 & \textbf{0.885} & \textbf{0.490} & 0.473 & \textbf{3.12} & 2.81 & \textbf{3.08} & 2.91 \\
    Gemini 2.0  & \textbf{0.991} & 0.874 & 0.403 & 0.439 & 2.96 & \textbf{3.09} & 2.91 & \textbf{3.22} \\
    \bottomrule
  \end{tabular}
  }%
\end{table}

\section{Dataset Details}
\label{sec:dataset_details}

\begin{figure*}[h]
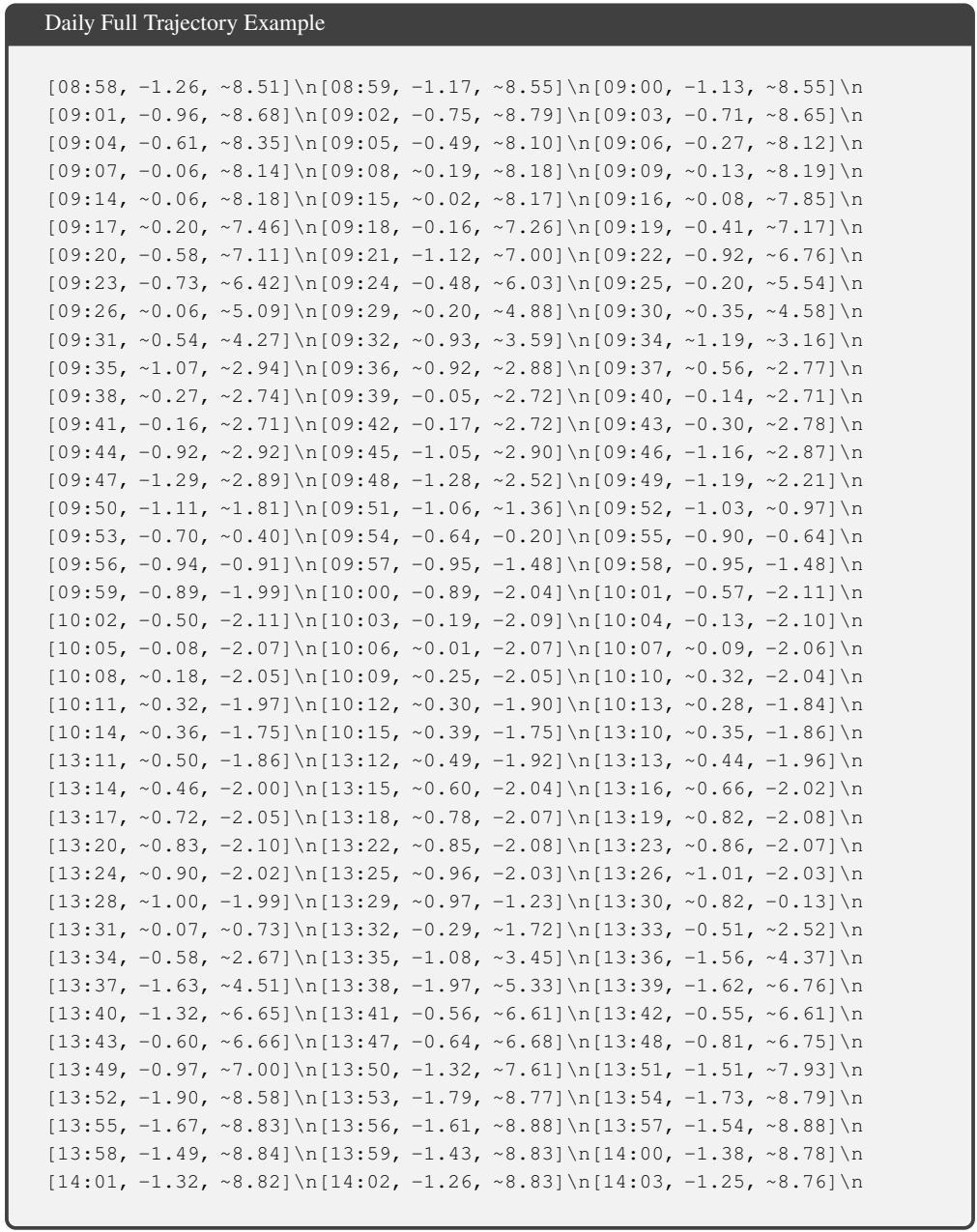

    \centering
    \begin{tcolorbox}[title=Daily Full Trajectory Example, width=0.75\textwidth, halign=center]
    \footnotesize
        \begin{lstlisting}[breaklines=true,basicstyle=\small\ttfamily, columns=flexible]
[08:58, -1.26, ~8.51]\n[08:59, -1.17, ~8.55]\n[09:00, -1.13, ~8.55]\n
[09:01, -0.96, ~8.68]\n[09:02, -0.75, ~8.79]\n[09:03, -0.71, ~8.65]\n
[09:04, -0.61, ~8.35]\n[09:05, -0.49, ~8.10]\n[09:06, -0.27, ~8.12]\n
[09:07, -0.06, ~8.14]\n[09:08, ~0.19, ~8.18]\n[09:09, ~0.13, ~8.19]\n
[09:14, ~0.06, ~8.18]\n[09:15, ~0.02, ~8.17]\n[09:16, ~0.08, ~7.85]\n
[09:17, ~0.20, ~7.46]\n[09:18, -0.16, ~7.26]\n[09:19, -0.41, ~7.17]\n
[09:20, -0.58, ~7.11]\n[09:21, -1.12, ~7.00]\n[09:22, -0.92, ~6.76]\n
[09:23, -0.73, ~6.42]\n[09:24, -0.48, ~6.03]\n[09:25, -0.20, ~5.54]\n
[09:26, ~0.06, ~5.09]\n[09:29, ~0.20, ~4.88]\n[09:30, ~0.35, ~4.58]\n
[09:31, ~0.54, ~4.27]\n[09:32, ~0.93, ~3.59]\n[09:34, ~1.19, ~3.16]\n
[09:35, ~1.07, ~2.94]\n[09:36, ~0.92, ~2.88]\n[09:37, ~0.56, ~2.77]\n
[09:38, ~0.27, ~2.74]\n[09:39, -0.05, ~2.72]\n[09:40, -0.14, ~2.71]\n
[09:41, -0.16, ~2.71]\n[09:42, -0.17, ~2.72]\n[09:43, -0.30, ~2.78]\n
[09:44, -0.92, ~2.92]\n[09:45, -1.05, ~2.90]\n[09:46, -1.16, ~2.87]\n
[09:47, -1.29, ~2.89]\n[09:48, -1.28, ~2.52]\n[09:49, -1.19, ~2.21]\n
[09:50, -1.11, ~1.81]\n[09:51, -1.06, ~1.36]\n[09:52, -1.03, ~0.97]\n
[09:53, -0.70, ~0.40]\n[09:54, -0.64, -0.20]\n[09:55, -0.90, -0.64]\n
[09:56, -0.94, -0.91]\n[09:57, -0.95, -1.48]\n[09:58, -0.95, -1.48]\n
[09:59, -0.89, -1.99]\n[10:00, -0.89, -2.04]\n[10:01, -0.57, -2.11]\n
[10:02, -0.50, -2.11]\n[10:03, -0.19, -2.09]\n[10:04, -0.13, -2.10]\n
[10:05, -0.08, -2.07]\n[10:06, ~0.01, -2.07]\n[10:07, ~0.09, -2.06]\n
[10:08, ~0.18, -2.05]\n[10:09, ~0.25, -2.05]\n[10:10, ~0.32, -2.04]\n
[10:11, ~0.32, -1.97]\n[10:12, ~0.30, -1.90]\n[10:13, ~0.28, -1.84]\n
[10:14, ~0.36, -1.75]\n[10:15, ~0.39, -1.75]\n[13:10, ~0.35, -1.86]\n
[13:11, ~0.50, -1.86]\n[13:12, ~0.49, -1.92]\n[13:13, ~0.44, -1.96]\n
[13:14, ~0.46, -2.00]\n[13:15, ~0.60, -2.04]\n[13:16, ~0.66, -2.02]\n
[13:17, ~0.72, -2.05]\n[13:18, ~0.78, -2.07]\n[13:19, ~0.82, -2.08]\n
[13:20, ~0.83, -2.10]\n[13:22, ~0.85, -2.08]\n[13:23, ~0.86, -2.07]\n
[13:24, ~0.90, -2.02]\n[13:25, ~0.96, -2.03]\n[13:26, ~1.01, -2.03]\n
[13:28, ~1.00, -1.99]\n[13:29, ~0.97, -1.23]\n[13:30, ~0.82, -0.13]\n
[13:31, ~0.07, ~0.73]\n[13:32, -0.29, ~1.72]\n[13:33, -0.51, ~2.52]\n
[13:34, -0.58, ~2.67]\n[13:35, -1.08, ~3.45]\n[13:36, -1.56, ~4.37]\n
[13:37, -1.63, ~4.51]\n[13:38, -1.97, ~5.33]\n[13:39, -1.62, ~6.76]\n
[13:40, -1.32, ~6.65]\n[13:41, -0.56, ~6.61]\n[13:42, -0.55, ~6.61]\n
[13:43, -0.60, ~6.66]\n[13:47, -0.64, ~6.68]\n[13:48, -0.81, ~6.75]\n
[13:49, -0.97, ~7.00]\n[13:50, -1.32, ~7.61]\n[13:51, -1.51, ~7.93]\n
[13:52, -1.90, ~8.58]\n[13:53, -1.79, ~8.77]\n[13:54, -1.73, ~8.79]\n
[13:55, -1.67, ~8.83]\n[13:56, -1.61, ~8.88]\n[13:57, -1.54, ~8.88]\n
[13:58, -1.49, ~8.84]\n[13:59, -1.43, ~8.83]\n[14:00, -1.38, ~8.78]\n
[14:01, -1.32, ~8.82]\n[14:02, -1.26, ~8.83]\n[14:03, -1.25, ~8.76]\n
        \end{lstlisting}
    \end{tcolorbox}
    \caption{Example of a full day trajectory in the MobQA dataset, showing timestamped coordinate points in the format [time, x-coordinate, y-coordinate]. The trajectory captures movement patterns from morning (08:58) to afternoon (14:03) with a gap between 10:15 and 13:10.}
    \label{fig:trajectory_example}
\end{figure*}

Tab.~\ref{tab:templates} presents the templates used for generating questions across all three task categories. Fig.~\ref{fig:trajectory_example} shows an example of a daily trajectory data, illustrating the format and structure of the textual trajectories in our dataset.

\begin{table*}[h!]
\centering
\caption{Description of templates used for question generation.}
\label{tab:templates}
\small
\begin{tabular}{lp{0.6\textwidth}}
\toprule
\textbf{Template Name} & \textbf{Template String} \\
\midrule
\multicolumn{2}{c}{\textbf{Factual Retrieval Question Templates}} \\
\midrule
\texttt{time\_to\_place} & \texttt{"Where was the person at \{time\}?"} \\
\texttt{place\_to\_time} & \texttt{"When was the person at \{place\}?"} \\
\texttt{frequency} & \texttt{"How many times did the person visit the location exactly at coordinates \{place\}?"} \\
\texttt{time\_to\_most\_frequent\_place} & \texttt{"Over the entire week, which location did the person visit most frequently during \{start time\}-\{end\_time\}?"} \\
\midrule
\multicolumn{2}{c}{\textbf{Multiple-Choice Question Templates}} \\
\midrule
\texttt{poi} & \texttt{Given that the person was at \{place\} at \{time\}, which point of interest did they most likely visit?} \\
\texttt{reason} & \texttt{Given that the person was at \{place\} at \{time\}, what was the most likely purpose of their visit?} \\
\texttt{mobility\_mode} & \texttt{The person was at \{place\} at \{time\}, what is the most likely mode of transportation?} \\
\texttt{next\_location} & \texttt{Given the trajectory up to \{time\}, where was the person most likely to go at \{next\_time\}?} \\
\texttt{time\_location\_prediction} & \texttt{Given the trajectory up to \{time\}, where was the person most likely to be at \{pred\_time\}?} \\
\texttt{place\_to\_timing\_prediction} & \texttt{Given the trajectory up to \{time\}, infer the time the person started their stay at \{place\} after \{time\}.} \\
\midrule
\multicolumn{2}{c}{\textbf{Free-Form Question Templates}} \\
\midrule
\texttt{routine} & \texttt{Analyze the weekly trajectory to identify and summarize the individual's regular movement patterns and routine activities} \\
\texttt{difference} & \texttt{Compare the person's movement patterns on \{day1\} and \{day2\}. How do they differ or are they similar?} \\
\texttt{anomaly} & \texttt{Evaluate whether the mobility pattern on \{day\} deviates from the individual's established routine (based on the rest of the week's data). Please explain the specific anomalies observed.} \\
\texttt{next\_location} & \texttt{Based on the trajectory up to \{time\}, predict where the person is most likely to go at \{next\_time\}.} \\
\texttt{time\_location\_prediction} & \texttt{Using the trajectory up to \{time\}, predict where the person is most likely to be at \{pred\_time\}.} \\
\texttt{place\_to\_timing\_prediction} & \texttt{Given the trajectory up to \{time\}, infer the time the person started their stay at \{place\} after \{time\}.} \\
\texttt{night\_anomaly} & \texttt{Identify any locations that were visited exclusively during late-night hours and discuss possible reasons for these visits.} \\
\texttt{occupation} & \texttt{Analyze the movement patterns and time data to infer where the person's home is located.} \\
\texttt{week\_social} & \texttt{Based on the trajectory data for the night of \{day\}, is there evidence that the person met with others or attended a social event? Please provide a explanation and answer.} \\
\texttt{week\_purpose} & \texttt{Given that the person is at \{place\} at \{time\}, and considering common weekly routines, what is the most likely activity taking place and the reason for being there? Please provide your reasoning.} \\
\texttt{day\_description} & \texttt{Analyze and describe the person's daily activities and events throughout the day.} \\
\texttt{social} & \texttt{Based on the given trajectory data, is there any evidence that the person met with others or attended a social event during the evening? Please provide a explanation and your answer.} \\
\bottomrule
\end{tabular}
\end{table*}

\section{Free-Form QA Examples}
\label{sec:free_form_examples}
Tab.~\ref{tab:case_study_traj_1} (Case Study 1) highlights significant differences in how the models interpret trajectories. As shown in the figure, this example uses a short daily trajectory for the QA task. The reference answer states that the coordinate (1.24, 0.58) at 22:07 lies on the commuter's home-bound bus route, \ie, a point passed through rather than stayed at. GPT-4o-mini completely overlooks this transit status, hallucinating a static activity (``dining at a restaurant or gathering with friends'') despite the unmistakable movement log. GPT-4o, while less erroneous, still mislabels the segment as a ``brief stop,'' failing to capture continued motion. On the other hand, o3-mini correctly recognizes that the person is in transit but misinfers the mobility mode, attributing the segment to walking instead of bus travel suggested by speed and trajectory context. In contrast, Gemini-1.5-Flash and Gemini-2.0-Flash both identify the segment as part of an ongoing journey home and explicitly cite linear coordinate progression and velocity changes as supporting evidence, achieving the highest faithfulness and informativeness scores.

Tab.~\ref{tab:case_study_traj_2} (Case Study 2) likewise highlights distinct strengths and weaknesses across the five models. GPT-4o-mini again retrieves only the most obvious surface‑level facts and therefore fails to deliver a substantive answer. GPT-4o and O3-mini correctly note that Day 7 spans a geographically wider area than Day 4, yet they do not extract richer qualitative insights such as activity purpose or temporal structure. In contrast, Gemini-1.5-Flash and Gemini-2.0-Flash ground their responses in objective trajectory evidence, but their outputs read largely as enumerations of coordinates and timestamps, offering little higher-level interpretation. Overall, none of the systems fully capture the nuanced, qualitative differences between the two days' mobility patterns.

These case studies illustrate the spectrum of capabilities among current LLMs in interpreting mobility data. While some models demonstrate proficiency in factual grounding, significant challenges remain in achieving nuanced semantic understanding, including accurate activity and mode recognition, and the generation of rich, qualitative interpretations of complex movement behaviors.

\begin{table*}[h]
    \centering
    \begin{threeparttable}
    \small
      \caption{Case study 1: model predictions and evaluation for the trajectory}
      \label{tab:case_study_traj_1}
      \begin{tabularx}{\linewidth}{lXcc}
        \toprule
        \multicolumn{4}{c}{\includegraphics[width=0.5\linewidth]{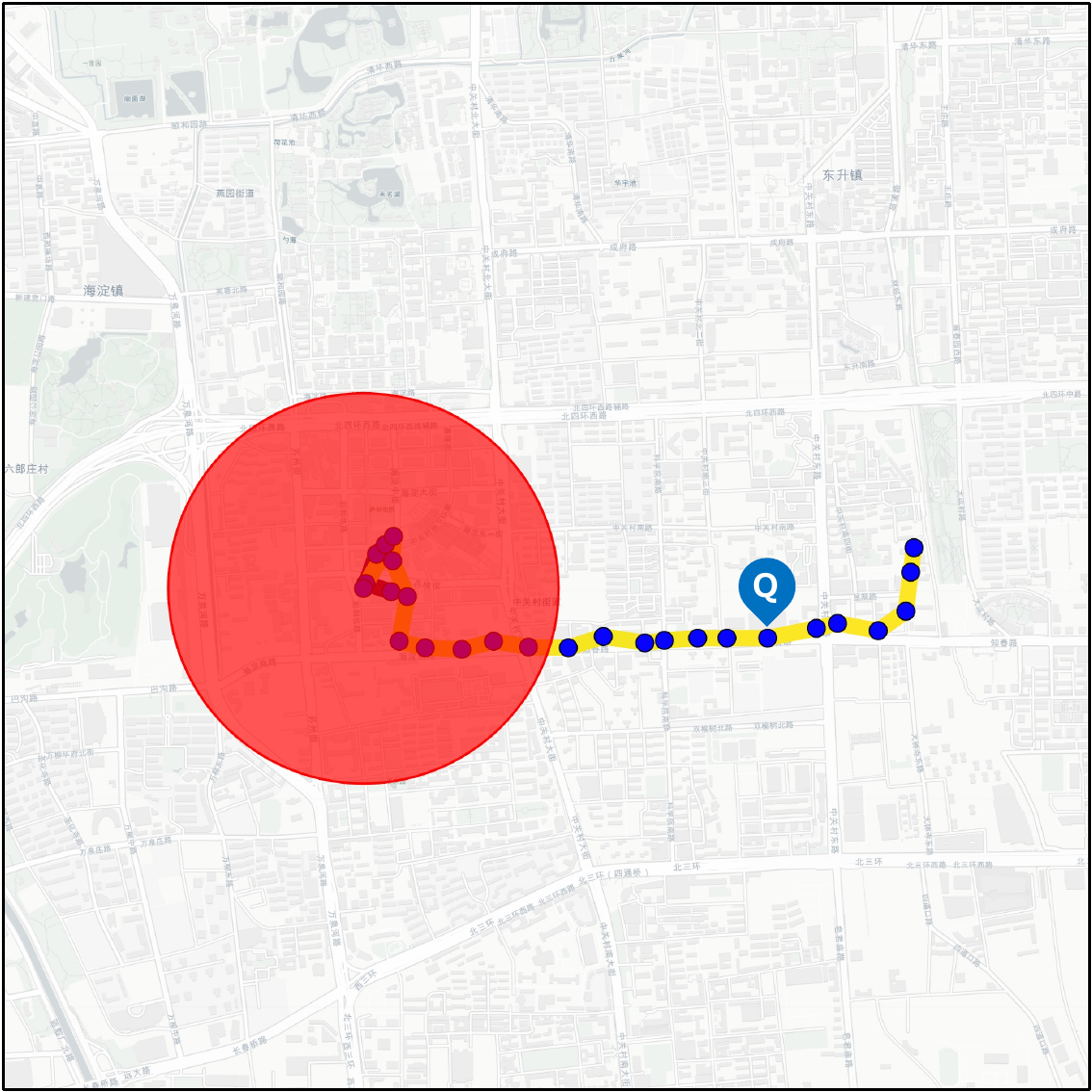}}\\
        \addlinespace
        \multicolumn{4}{p{\linewidth}}{\textbf{Question}: Given that the person is at (1.24, 0.58) at 22:07, what is the most likely purpose of their visit? Please include both the expected activity and the underlying reason.}\\
        \addlinespace[0.3em]
        \multicolumn{4}{p{\linewidth}}{\textbf{Reference answer}: The reason for staying at position (1.24, 0.58) at 22:07 is likely "returning home" as this location is on a bus route during their commute back home.}\\
        \midrule
        \textbf{Model} & \textbf{Generated answer} & \textbf{Faith.}\tnote{a} & \textbf{Info.}\tnote{a}\\
        \midrule
        GPT-4o-mini      & Based on this location, the purpose of the visit is likely dining at a restaurant or gathering with friends. & 1 & 1\\
        GPT-4o           & Given that they were at (1.46, 0.62) at 22:08, the expected activity at this location is likely a brief stop or intermediate point during transit. & 2 & 2\\
        O3-mini          & The subject is likely out for an evening walk or strolling. This can be inferred from their walking speed and continuous movement pattern, suggesting regular pedestrian activity aimed at relaxation or exercise. & 4 & 3\\
        Gemini-1.5-Flash & The purpose of the person's visit at (1.46, 0.62) at 22:08 is most likely transit to an evening destination. The data shows nearly linear movement from (-0.45, 1.04) to (1.90, 0.98) between 21:53 and 22:13, which can be interpreted as a direct route to a destination. & 5 & 4\\
        Gemini-2.0-Flash & The person at (1.46, 0.62) at 22:08 has been moving in one direction since 22:03 and will be at (1.90, 0.98) by 22:13, suggesting they are simply passing through this location rather than visiting it for a specific purpose. & 5 & 4\\
        \bottomrule
      \end{tabularx}
      \begin{tablenotes}
        \item [a] Scores are on a 1-5 ordinal scale: higher is better in informativeness and faithfulness.
      \end{tablenotes}
    \end{threeparttable}
  \end{table*}

\begin{table*}[h]
    \centering
    \begin{threeparttable}
    \small
      \caption{Case study 2: model predictions and evaluation for the trajectory}
      \label{tab:case_study_traj_2}
      \begin{tabularx}{\linewidth}{lXcc}
        \toprule
        \multicolumn{4}{c}{\includegraphics[width=1\linewidth]{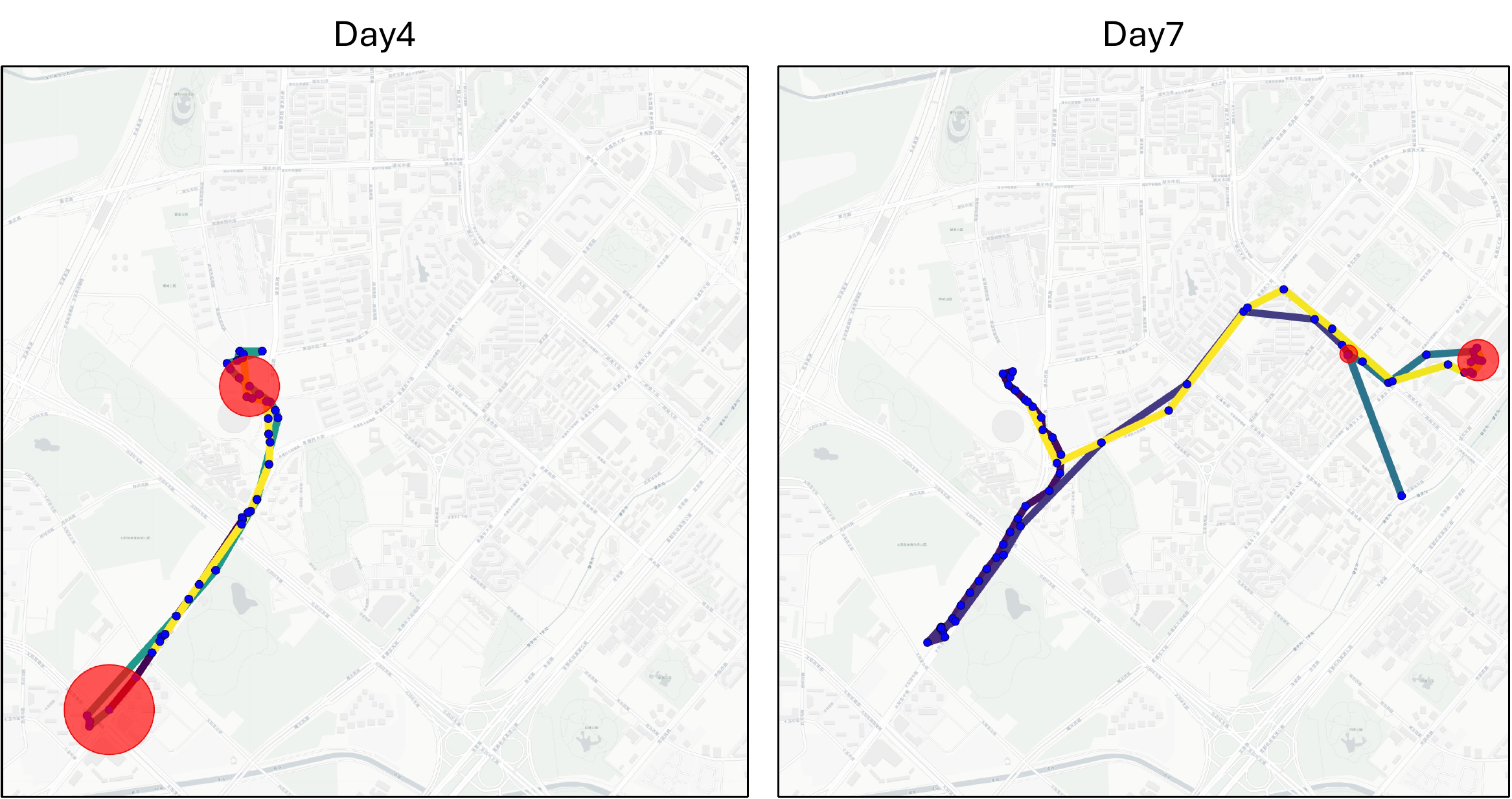}}\\
        \addlinespace
        \multicolumn{4}{p{\linewidth}}{\textbf{Question}: Compare the person’s movement patterns on Day 4 and Day 7. How do they differ or are they similar?}\\
        \addlinespace[0.3em]
        \multicolumn{4}{p{\linewidth}}{\textbf{Reference answer}: On Day 4, there is primarily a long stay at the workplace, with the main activity being commuting between home and work. In contrast, Day 7 involves movement to multiple destinations, including a shopping center. While Day 4 is characterized by long stays at the workplace, Day 7 is distinguished by multi-purpose visits, centered around continuous short-term movements including time spent at a shopping center.}\\
        \midrule
        \textbf{Model} & \textbf{Generated answer} & \textbf{Faith.}\tnote{a} & \textbf{Info.}\tnote{a}\\
        \midrule
        GPT-4o-mini      & The movement patterns on Day 4 and Day 7 are very different, as on Day 4 the person spent most of their time at the same location, while on Day 7 they were actively moving around. & 2 & 2\\
        GPT-4o           & Day 4 focuses primarily on local movements, whereas Day 7 shows travel across a wider area, indicating more active exploration. The days differ in terms of movement range and destination stability. & 3 & 2\\
        O3-mini          & Day 4 is characterized by short-distance movements within the same small area, showing localized activity, while Day 7 clearly shows movement across a wider range from morning to noon, covering greater distances. & 3 & 3\\
        Gemini-1.5-Flash & On Day 4, the person remained in almost the same location from around midnight until about 6:50 AM, then after a brief movement from 12:50 to around 1:16 PM, they returned to the same location. In contrast, on Day 7, they moved within a relatively narrow range from 5:49 AM to around 6:44 AM, then paused briefly before engaging in long, extensive movement from 7:35 AM to around 10:21 AM. Therefore, Day 4 is characterized by relatively stationary periods, while Day 7 features long-distance travel. & 4 & 2\\
        Gemini-2.0-Flash & Comparing the movements on Days 4 and 7, on Day 4 the person moved from around (-0.8, -0.9)km to around (0.1, 0.8)km from late night to early morning and after noon, then tended to stay near (0.1, 0.8)km. On Day 7, they moved from around (-0.1, 0.8)km to around (2.2, 0.9)km in the early morning, spent most of the morning near (2.2, 0.9)km, then moved from around (2.2, 0.9)km to around (0.1, 0.7)km. In other words, on Day 4 they spent more time staying near (0.1, 0.8)km, while on Day 7 they spent more time near (2.2, 0.9)km. Additionally, the travel distance on Day 7 was longer than on Day 4. & 5 & 3\\
        \bottomrule
      \end{tabularx}
      \begin{tablenotes}
        \item [a] Scores are on a 1–5 ordinal scale: higher is better in informativeness and faithfulness.
      \end{tablenotes}
    \end{threeparttable}
  \end{table*}

\end{document}